\documentclass{article} % For LaTeX2e
\usepackage{collas2024_conference,times}
\usepackage{easyReview}

\usepackage{graphicx}
\usepackage{amsmath}
\usepackage{amssymb}
\usepackage{booktabs}
\usepackage{xspace}
\usepackage{subcaption}
\usepackage{caption}

\usepackage{times}
\usepackage{epsfig}
\usepackage{float}
\usepackage{xfrac}
\usepackage{bm}
\usepackage{enumitem}
\usepackage{pifont}
\usepackage{color, colortbl}
\usepackage{wrapfig}

\definecolor{JungleGreen}{rgb}{0.16, 0.67, 0.53}
\definecolor{RedViolet}{rgb}{0.78, 0.08, 0.52}
\definecolor{Red}{rgb}{1.0, 0.0, 0.0}
\definecolor{Green}{rgb}{0.0, 0.5, 0.0}
\definecolor{DarkGray}{rgb}{0.41, 0.41, 0.41}
	
\definecolor{Gray}{gray}{0.9}

\newcommand{\COCOtoLVIS}{COCO$\rightarrow$LVIS\xspace}
\newcommand{\basecolor}[1]{\textcolor{JungleGreen}{\textbf{#1}}}
\newcommand{\novelcolor}[1]{\textcolor{RedViolet}{\textbf{#1}}}
\newcommand{\myparagraph}[1]{\noindent\textbf{#1}.}
\newcommand{\COCOhalf}{$\text{COCO}_{half}$}

\newcommand{\beginsupplement}{%
        \setcounter{table}{0}
        \renewcommand{\thetable}{A\arabic{table}}%
        
        \setcounter{figure}{0}
        \renewcommand{\thefigure}{A\arabic{figure}}%
    
        \setcounter{equation}{0}
        \renewcommand{\theequation}{A\arabic{equation}}%
        
        \setcounter{section}{0}
        \renewcommand{\thesection}{A\arabic{section}}%
        
        \setcounter{page}{1}
        \renewcommand{\thepage}{A\arabic{page}}%
}

\usepackage{amsmath,amsfonts,bm}

\def\eqref#1{equation~\ref{#1}}
\def\1{\bm{1}}

\DeclareMathAlphabet{\mathsfit}{\encodingdefault}{\sfdefault}{m}{sl}
\SetMathAlphabet{\mathsfit}{bold}{\encodingdefault}{\sfdefault}{bx}{n}

\usepackage{hyperref}
\hypersetup{
    colorlinks=true,
    linkcolor=red,
    filecolor=magenta,
    urlcolor=blue,
    citecolor=purple,
    pdftitle={PANDAS},
    pdfpagemode=FullScreen,
}

\title{PANDAS: Prototype-based\\Novel Class Discovery and Detection}

\author{Tyler~L.~Hayes, C\'{e}sar~R.~de~Souza, Namil Kim, Jiwon Kim, Riccardo Volpi, Diane Larlus\\
NAVER LABS\\
}

\collasfinalcopy % Uncomment for camera-ready version, but NOT for submission.

\begin{document}

\maketitle

\begin{abstract}
Object detectors are typically trained once and for all on a fixed set of classes. However, this closed-world assumption is unrealistic in practice, as new classes will inevitably emerge after the detector is deployed in the wild. In this work, we look at ways to extend a detector trained for a set of base classes so it can i) spot the presence of novel classes, and ii) automatically enrich its repertoire to be able to detect those newly discovered classes together with the base ones. We propose PANDAS, a method for novel class discovery and detection. It discovers clusters representing novel classes from unlabeled data, and represents old and new classes with prototypes. During inference, a distance-based classifier uses these prototypes to assign a label to each detected object instance. The simplicity of our method makes it widely applicable. We experimentally demonstrate the effectiveness of PANDAS on the VOC 2012 and COCO-to-LVIS benchmarks. It performs favorably against the state of the art for this task while being computationally more affordable.\footnote{Code is available at: \url{https://github.com/naver/pandas}}
\end{abstract}

% ----------------------------------------------------
\section{Introduction}
\label{sec:intro}

Object detection is a fundamental computer vision task that aims at localizing and classifying any object instance from a predefined list of object categories. This requires the detector to be provided up-front with a set of images annotated for the categories that it should learn to recognize and localize. Yet, in real-world applications, some object categories are not seen or annotated during training and could turn out to be important to detect.

Several paradigms seek to address this problem, which mostly differ in the level of manual intervention they require. For example, the detector could be enhanced with additional categories by gathering new data or annotations, and updating it via retraining or fine-tuning. This is how class incremental detection operates~\citep{cermelli2022incremental}.
In open-world learning~\citep{bendale2015towards,joseph2021towards}, detectors are equipped with a mechanism that allows them to recognize when a detected object is unknown. Yet, these methods still require manual annotations to extend the detector once the unknown objects have been identified. Moreover, in some cases, gathering annotations for retraining is not possible, and enabling a detector to handle novel classes must happen in a fully automated way. This is the more challenging scenario that we consider in this paper, which is often referred to as \textit{novel class discovery}~\citep{han2019learning,vaze2022generalized,troisemaine2023novel}.

Novel Class Discovery (NCD) is challenging for several reasons. First, similar to open-set~\citep{scheirer2012toward} or open-world learning~\citep{bendale2015towards}, it requires distinguishing potentially relevant novel classes from what are merely unseen background patterns. Second, it requires the model to maintain performance on the base classes, since adding novel classes could lead to catastrophic forgetting of the base ones~\citep{mccloskey1989}. 
Finally, novel classes should be discovered and added to the pool of classes the model can detect without manual intervention.

While novel class discovery has been abundantly studied for classification (see \citet{troisemaine2023novel} for a survey), it has been seldom considered for detection~\citep{Fomenko_2022_NeurIPS,Wu_2022_ECCV,Zheng_2022_CVPR,rambhatla2021pursuit}, despite being of great practical use. Departing from existing work, we propose a simple yet effective method that i) does not require storing base images during the novel class discovery phase and ii) does not require knowing the number of novel classes a priori. We call our method \textbf{PANDAS}, which stands for \textbf{P}rototype b\textbf{A}sed \textbf{N}ovel class \textbf{D}etection \textbf{A}nd di\textbf{S}covery. First, it computes prototypes for labeled base classes. Then, it learns clusters and their associated prototypes to represent novel classes from unlabeled data during a discovery phase. Base and novel prototypes are used in conjunction with a simple distance-based classifier that allows detecting both base and discovered novel classes at test time. In a thorough experimental analysis, we show that PANDAS is competitive with the state-of-the-art RNCDL method~\citep{Fomenko_2022_NeurIPS}, despite its simplicity and more affordable compute cost. For example, on VOC, PANDAS outperforms RNCDL by 12.0\% mAP on novel classes while running over 36 times faster. On LVIS, PANDAS outperforms RNCDL by 2.5\% mAP on novel classes while running over five times faster.

To summarize, we contribute to the challenging task of novel class discovery and detection and propose PANDAS, a novel method that is simple, scalable, and does not require privileged information such as the number of classes to discover. We empirically show that PANDAS outperforms the state of the art~\citep{Fomenko_2022_NeurIPS} on PASCAL VOC 2012 and the more challenging COCO-to-LVIS benchmark, while being much simpler and running significantly faster. Finally, we show qualitative visualizations providing evidence that PANDAS can sometimes detect objects not present in the ground truth, showing improvements beyond what is captured by quantitative NCD metrics alone. 

% ----------------------------------------------------
\begin{figure}[t!]
    \centering
    \includegraphics[width=0.75\linewidth]{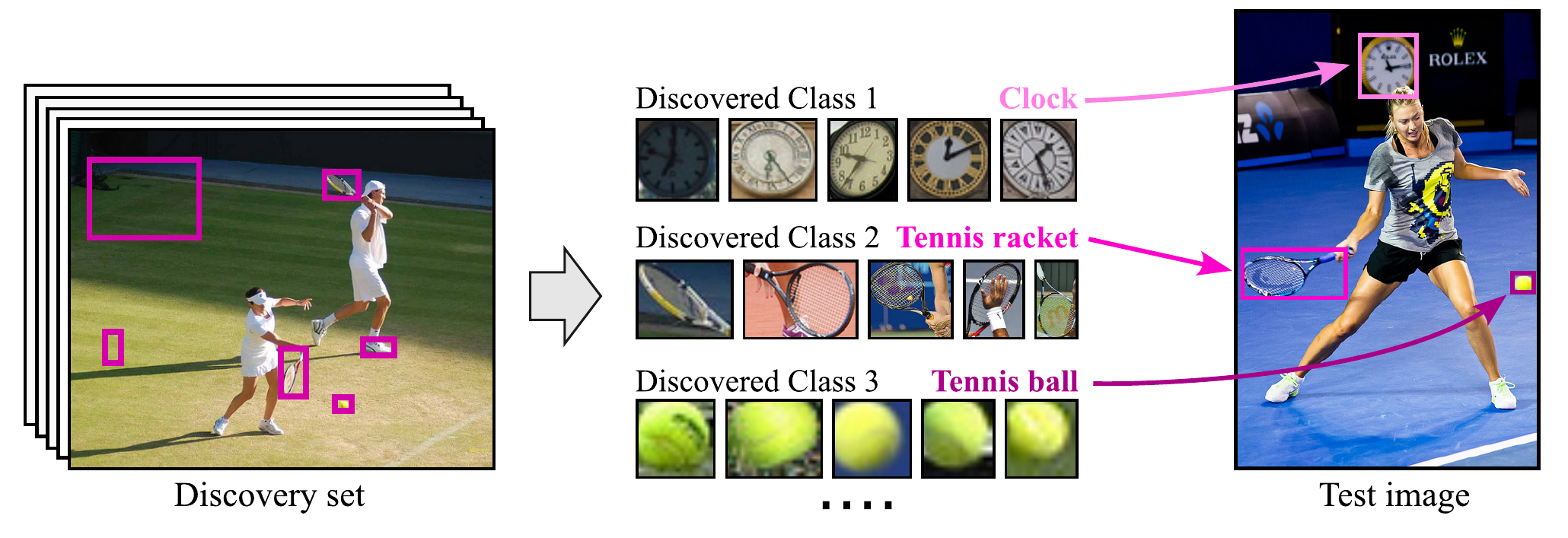}
    \caption{
    This paper tackles the task of \textit{novel class discovery and detection}. We start from an initial detector, trained for base classes, and aim to extend it by automatically discovering \novelcolor{novel} classes from a set of unannotated images (i.e., the Discovery set). \novelcolor{Novel} classes are added to the detector, on the fly, without any manual intervention. The updated detector must then detect both base classes and \novelcolor{novel} classes in new test images. Note that this example only shows \novelcolor{novel} discovered classes for clarity.
    }
    \label{fig:enter-label}
\end{figure}
% ----------------------------------------------------

% ----------------------------------------------------

\section{Related Works}
\label{sec:related-works}

\myparagraph{Novel class discovery}~Classification methods targeting Novel Class Discovery (NCD) are provided training data containing labeled examples from some semantic classes and unlabeled examples from other classes. The goal of NCD methods is to then learn to group examples from novel classes together and identify when a new example belongs to one of the novel groups (see \citet{troisemaine2023novel} for a survey). Some of the earliest classification techniques for NCD used supervised data to learn priors for grouping samples together and then transferred the learned priors via similarity functions to unlabeled data~\citep{hsu2016deep,hsu2018learning}. \citet{han2019learning} proposed a two-stage pipeline for the NCD problem. In the pipeline, generalized feature representations are learned using labeled data, and then clustering is applied to the unlabeled data by transferring features from the first (supervised) stage. Specifically, \citet{han2019learning} used deep transfer clustering to cluster unknowns, and proposed a technique to estimate the number of classes in the unlabeled dataset. Later work built on this two-stage framework by learning features via self-supervised learning and using ranking statistics to determine the similarity between unlabeled examples~\citep{Han2020Automatically}. Many works have built on this two-stage pipeline and improve performance via self-supervised feature learning and regularization~\citep{jia2021joint,zhao2021novel,zhong2021neighborhood,zhong2021openmix}. Instead of using separate losses for labeled and unlabeled data, the UNified Objective (UNO) method produces pseudo-labels for unlabeled examples via multi-view synthesis and updates a network with a single objective~\citep{fini2021unified}. More recently, NCD methods have been extended to predict known and novel classes in test images~\citep{cao2022openworld,vaze2022generalized}; this is sometimes referred to as \emph{Generalized Category Discovery}~\citep{vaze2022generalized}.

\myparagraph{Novel class discovery for detection}~While NCD has been extensively studied for classification (see \citet{troisemaine2023novel} for a survey), there are few works on NCD for object detection~\citep{Fomenko_2022_NeurIPS,Wu_2022_ECCV,Zheng_2022_CVPR,rambhatla2021pursuit} and even fewer works on NCD for semantic segmentation~\citep{zhao2022novel}. Although detection requires the classification of objects from known and unknown classes, it also requires additional capabilities including localizing objects within an image (regression) and managing their co-occurrence. This is in contrast to many classification problems where an image contains a single object of interest. The Region-based Novel Class discovery, Detection, and Localization (RNCDL) method~\citep{Fomenko_2022_NeurIPS} first learns the parameters of an R-CNN network with a class-agnostic regression head using supervised learning. Then, it updates the known-class classification head jointly with a novel-class classification head using the Swapping Assignments between Views (SwAV) self-supervised learning method~\citep{caron2020unsupervised} on generated pseudo-labels, which are assumed to have a long-tailed distribution. However, self-supervised methods often require long training times and consume large amounts of resources. Moreover, the distribution of labels in the wild is usually unknown. \citet{Zheng_2022_CVPR} first train an open-set object detector to distinguish known classes from unknowns. 
Then, an unsupervised contrastive loss is used to learn more discriminative representations, before using a constrained version of $k$-means clustering to cluster novel classes. In \citet{Wu_2022_ECCV}, a pseudo-labeling process based on objectness scores is used to train a known versus unknown classification head. Then, a similarity loss is used to assign novel classes to clusters, which are then improved via cluster refinement. \citet{rambhatla2021pursuit} use the feature space of an R-CNN to maintain two memory modules for discovering novel classes. Specifically, the semantic memory uses Linear Discriminant Analysis (LDA) classifiers to determine if an object is known or unknown, and if it is unknown, then a cluster in the working memory is updated with the current object feature. Our method is simpler than most existing methods as it does not require knowing the number of classes to discover~\citep{Wu_2022_ECCV}, does not need to store images from base classes to prevent catastrophic forgetting~\citep{Fomenko_2022_NeurIPS,Zheng_2022_CVPR}, and does not require a class prior on the novel class distribution~\citep{Fomenko_2022_NeurIPS}.

\myparagraph{Related paradigms}~There are several related and complementary research fields to NCD, which we briefly discuss and contrast with NCD here. In \emph{open-set learning}, the test set is considered \emph{open}, where examples can come from classes inside or outside of the training distribution~\citep{scheirer2012toward,jain2014multi,scheirer2014probability,bendale2016towards}. Open-set methods are then required to identify if a test sample is from a known class or unknown, and has been studied for detection by \citet{miller2018dropout,dhamija2020overlooked}. \citet{bendale2015towards} extend the open-set learning paradigm to \emph{open-world learning}, which allows models to incrementally update on samples from unknown classes over time, after the unknowns have been manually annotated. Different from NCD, open-set and open-world learning only require methods to identify if a sample is ``unknown'', while NCD requires methods to additionally group unknown samples from the same classes together. Open-world learning has recently been studied for object detection~\citep{joseph2021towards,gupta2022owdetr}. In the work of \citet{joseph2021towards}, a detector was trained with contrastive clustering and then an energy-based model was used to detect unknowns. The effectiveness of transformers in open-world detection has also been studied~\citep{gupta2022owdetr}.

Beyond open-set and open-world learning, NCD is also related to fully unsupervised paradigms. \emph{Unsupervised object discovery and localization} methods are required to localize objects in images from an unlabeled pool of images~\citep{tuytelaars2010unsupervised,cho2015unsupervised,lee2010object,lee2011learning,vo2021large,rambhatla2023most,wang2022freesolo} (see \citet{simeoni2023unsupervised} for a survey of the most recent methods). In contrast to NCD, unsupervised object discovery methods do not receive labeled data during training and are not required to group knowns and unknowns together. \emph{Zero-shot learning} methods use additional meta-data such as attribute vectors or latent vectors to identify unknowns~\citep{xian2018zero,bansal2018zero}. Similarly, \emph{open-vocabulary} methods use auxiliary data from large vision and language models such as text embeddings to train models~\citep{zhou2022detecting}. In contrast, NCD methods do not assume access to any auxiliary data about unknown classes during their discovery phase.

% ----------------------------------------------------

\section{Problem Formulation}
\label{sec:problem-formulation}
We study the problem of \textit{novel class discovery and detection} introduced by \citet{Fomenko_2022_NeurIPS}, which requires a detector to first learn from labeled base images and then transfer the learned base knowledge to discover novel classes in an unsupervised way. More specifically, a method first learns the parameters of a detection model during a \emph{Base Phase} (Sec.~\ref{subsec:base}) from a dataset $\mathcal{D}^{k}$ labeled for base (known) classes with label set $\mathcal{C}^{k}$ such that $\left| \mathcal{C}^{k}\right|=K$. We define $\mathcal{D}^{k} = \left\{\left(\mathbf{I},\left\{(c,x,y,w,h)\right\}\right)\right\}$, where $\mathbf{I}$ is an image with its associated set of object instances, each defined by the object label $c \in \mathcal{C}^{k}$ and bounding box $\left(x,y,w,h\right)$ where $x$ and $y$ correspond to the top-left coordinate, and $w$ and $h$ are the box width and height, respectively.
Then, during a \emph{Discovery Phase} (Sec.~\ref{subsec:discovery}), the method is provided with an unlabeled discovery dataset $D^{u}=\left\{\mathbf{I}\right\}$, possibly containing both base classes in $\mathcal{C}^{k}$ and novel (\textit{unknown}) classes in $\mathcal{C}^{u}$ such that $\left| \mathcal{C}^{u} \right|=P$ and $\mathcal{C}^{k} \bigcap \mathcal{C}^{u} = \emptyset$. After learning from the discovery dataset, the method should correctly detect all object instances in an arbitrary image by localizing them and predicting their associated label in $\mathcal{C}^{k}$ if they are base classes, or predicting the label corresponding to their discovered category otherwise (see Sec.~\ref{subsec:semantic-classes}).

% ----------------------------------------------------

\section{Method}
\label{sec:method}

\begin{figure*}[t!]
     \includegraphics[width=\linewidth]{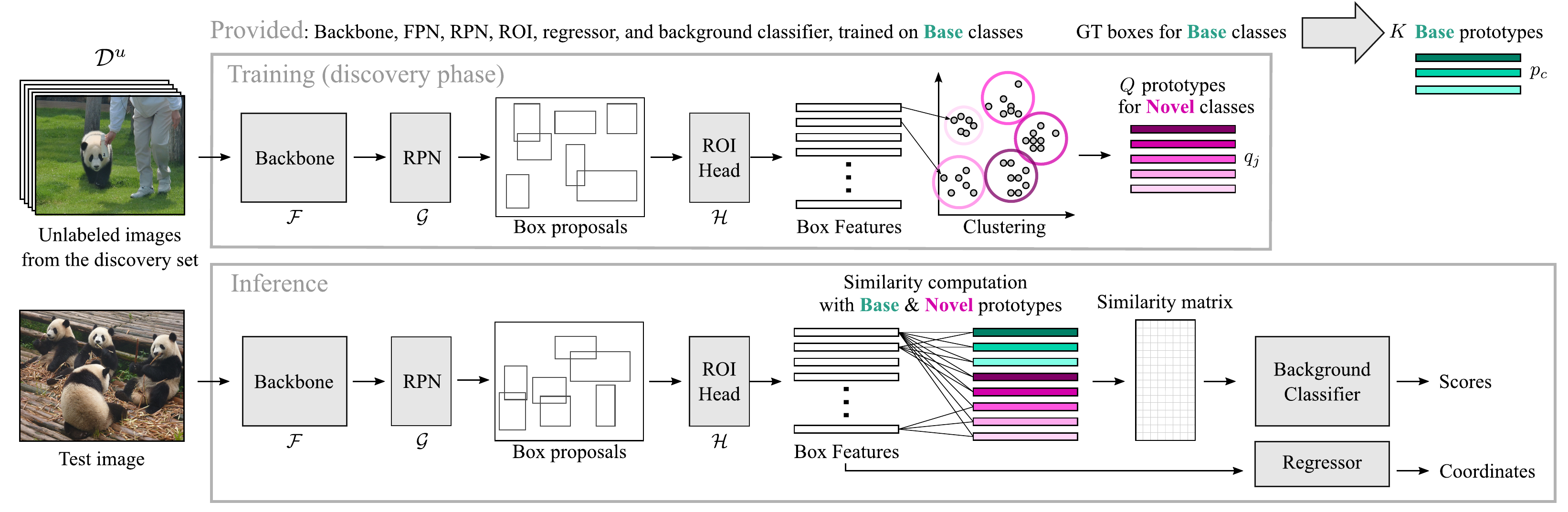}
     \caption{
     \textbf{Overview of our PANDAS method.} Given an annotated training set and a detector trained for \basecolor{Base} classes on that training set, we compute \basecolor{Base} prototypes. Then components from this detector are used during the discovery phase to compute prototypes for the \novelcolor{Novel} classes using clustering (top). During inference, similarity scores of new box features to these prototypes are computed to provide predicted scores (bottom).}
     \label{fig:methods}
\end{figure*}

We outline our general pipeline, including the base (Sec.~\ref{subsec:base}) and discovery (Sec.~\ref{subsec:discovery}) phases. Then, we describe our inference procedure (Sec.~\ref{subsec:inference}), including how we assign semantic class names to discovered clusters (Sec.~\ref{subsec:semantic-classes}).

Following \citet{Fomenko_2022_NeurIPS}, our method builds on the Faster R-CNN detector~\citep{ren2015faster} with a Feature Pyramid Network (FPN)~\citep{lin2017feature} and ResNet-50 backbone~\citep{He_2016_CVPR}, $\mathcal{F}\left(\cdot\right)$. Faster R-CNN is a two-stage detector: it first selects candidate bounding boxes that are likely to contain an object and then further refines and classifies them. More precisely, in the first stage, a class-agnostic Region Proposal Network (RPN) $\mathcal{G}\left(\cdot\right)$ determines if candidate bounding 
boxes contain foreground objects or background regions, 
producing features for each object in an image as 
$\mathbf{b}_{i}=\mathcal{G}\left(\mathcal{F}\left(\mathbf{I}\right)\right)$, where $\mathbf{b}_{i}$ corresponds to the $i$-th box. In the second stage, the network pools each box feature $\mathbf{b}_{i}$ using Region of Interest (RoI) pooling and applies several fully-connected layers $\mathcal{H}\left(\cdot\right)$ to produce a more disentangled object feature, i.e., $\mathbf{f} = \mathcal{H}\left(\mathbf{b}_{i}\right)$. Finally, a classification head $\mathcal{O}_{c}\left(\cdot\right)$ and a regression head $\mathcal{O}_{r}\left(\cdot\right)$ are applied to the object feature to produce class predictions: $\hat{c} = \mathcal{O}_{c}\left(\mathbf{f}\right)$ and bounding box regression coordinates: $\hat{r} = \mathcal{O}_{r}\left(\mathbf{f}\right)$. Next, we discuss our modifications to the basic detection architecture for novel class detection.

% ----------------------------------------------------
\subsection{Base Phase}
\label{subsec:base}

During the base phase, the Faster R-CNN network learns to detect base class objects from labeled data. Traditionally, Faster R-CNN uses a per-class regression head to precisely localize individual classes. However, since we are interested in localizing labeled classes during the base phase and potentially unlabeled classes during the discovery phase, we replace this per-class regression head with a class-agnostic regression head, which outputs four regression coordinates (a bounding box) independent of the predicted class. Then, we train the network using standard supervised detection cross-entropy losses. After base training, we extract features from the RoI head, $\mathcal{H}$, for all ground truth boxes in the base training data $\mathcal{D}^{k}$ and compute one prototype per class. That is, given each ground truth box feature $\mathbf{f}^{GT}$ from class $c$, we compute the class prototype $\mathbf{p}_{c}$ as:
\begin{equation}
    \label{eq: gt prototypes}
    \mathbf{p}_{c} = \frac{1}{M}\sum_{i=1}^{M} \frac{\mathbf{f}_{i}^{GT}}{\lVert\mathbf{f}_{i}^{GT}\rVert_{2}} \enspace ,
\end{equation}
where $\lVert\cdot\rVert_{2}$ denotes the $L_{2}$ norm and $M$ is the number of samples for class $c$. We use these class prototypes along with their associated labels to classify base classes during discovery and evaluation. Note that our method stores only base class prototypes, and not the entire base dataset. This is in contrast to existing methods which require storing the entire base set, and is beneficial as it 
reduces memory requirements and potentially reduces privacy concerns.

% ----------------------------------------------------
\subsection{Discovery Phase}
\label{subsec:discovery}

During the discovery phase, the network should learn to detect novel classes and to distinguish them from one another, while maintaining the ability to detect base classes. We keep the Faster R-CNN weights from the base phase frozen during the discovery phase. Given a discovery dataset $\mathcal{D}^{u}$, we first use the network learned during the base phase to predict RPN box candidates for each image and extract their associated feature vectors $\mathbf{f}^{RPN}$. We then $L_{2}$-normalize these features and perform clustering on the feature vectors from all images in the discovery dataset. The resulting $Q$ cluster centers $\left\{\mathbf{q}_{j}\right\}_{j=1}^{Q}$ represent different groups that our method discovered, which we refer to as novel prototypes.

% ----------------------------------------------------
\subsection{Inference}
\label{subsec:inference}

During inference, our method is tasked with the detection and localization of both base and novel class objects in new images. Specifically, given a test image $\mathbf{I}$, our method predicts all RPN box candidates for this image and extracts the associated features $\mathbf{f}^{RPN}$. Next, we compute a logit vector for each feature, which will contain one score for the background class, $K$ scores for each of the $K$ base classes, and $Q$ scores for each of the $Q$ clusters.

To compute these logit vectors, we first compute the similarity of each $L_{2}$-normalized RPN feature, $\sfrac{\mathbf{f}^{RPN}}{\lVert\mathbf{f}^{RPN}\rVert_{2}}$, to each of the $K$ base prototypes and each of the $Q$ novel prototypes using similarity metric $\mathcal{S}$. We study several similarity metrics in Sec.~\ref{subsec:results-voc}. Next, to compute the background logit for each feature, we first compute the classification output of the original base classifier as $\hat{c} = \mathcal{O}_{c}\left(\mathbf{f}^{RPN}\right)$. If the predicted class $\hat{c}$ is ``background'', then we assign the background logit score for this feature as the maximum logit across all logits computed in the previous step. That is, if the base network classifies a region as ``background'', then PANDAS also classifies the region as ``background''. We call this mechanism the \emph{background classifier}.

After we compute logit vectors for each box, we then perform post-processing. We first convert the logit vectors into probability vectors that sum to one using either softmax or $L_{1}$-normalization, which we study in Sec.~\ref{subsec:results-voc}. Then, we follow the standard Faster R-CNN post-processing steps (i.e., remove low-scoring and empty boxes, perform non-maximal suppression, and keep the top-$m$ scoring predictions).

Finally, these classification predictions are output by the network, along with regression coordinates from the class-agnostic regressor from the original base model. That is, the regression coordinates are given by $\hat{r} = \mathcal{O}_{r}\left(\mathbf{f}^{RPN}\right)$.

% ----------------------------------------------------
\subsection{Assigning Semantic Classes to Clusters}
\label{subsec:semantic-classes}

Once novel clusters are ``discovered'', we need to assign semantic class labels to our computed clusters to assess the performance of our method. That is, we must map the novel class labels from the ground truth to our discovered clusters. To do this, we follow the evaluation protocol that is standard in NCD and that was introduced for detection by \citet{Fomenko_2022_NeurIPS}. First, we extract ground truth box features $\mathbf{f}^{GT}$ for all objects in the evaluation dataset with \emph{novel} class labels. We then compute the most similar cluster center to each of these features using the similarity metric $\mathcal{S}$. This provides a vector of ground truth labels and a vector of predicted cluster labels. We then run Hungarian matching~\citep{kuhn1955hungarian} on these two vectors to yield a mapping of ground truth labels to cluster labels. This mapping may not be one-to-one since the number of clusters may differ from the number of ground truth novel class labels. Finally, we run the inference procedure from Sec.~\ref{subsec:inference} on the evaluation dataset. If the network predicts a base class, we output the stored label for that base class. If the network predicts a cluster, we use the mapping from the Hungarian algorithm to output a label. Note that this mapping procedure only needs to be run once after a discovery phase (before evaluation). This means that inference can be performed in real-time since the mapping is pre-computed offline. Since it may not be possible to collect annotations of the novel classes in practical scenarios, we also explore the use of a vision-language model for mapping discovered clusters to semantic labels in Sec.~\ref{supp:timing}.

% ----------------------------------------------------
\section{Experimental Setup}
\label{sec:experiments}

% ----------------------------------------------------
\myparagraph{Benchmarks}~We use two detection benchmarks for NCD. The first one is based on Pascal VOC 2012 (VOC)~\citep{everingham2010pascal}. VOC is a popular object detection benchmark containing 10,582 training images and 1,449 validation images from 20 classes. We use the 10-10 split from \citet{cermelli2022incremental} that assigns 10 classes to the base classes and 10 to the novel ones (see Appendix Fig.~\ref{fig:voc-distributions} for the distributions). This 10-10 split contains a base set with images containing 10 classes and a discovery set with images potentially containing all 20 classes. We report performance on the full validation set, as well as on objects from base classes and novel classes separately.

The second NCD benchmark, \COCOtoLVIS is based on the MS COCO 2017~\citep{lin2014microsoft} and LVIS v1~\citep{gupta2019lvis} datasets and was proposed by \citet{Fomenko_2022_NeurIPS}. It is significantly more challenging than VOC, with 100,170 training images, 19,809 validation images, and 1,203 total classes that follow a long-tailed class distribution. Note that LVIS v1 and COCO use the same images, but have different training and validation sets. Specifically, the base training dataset contains half of the images from the COCO dataset with annotations for all 80 COCO classes. The discovery dataset then consists of the remaining images in the LVIS v1 dataset and 1,203 classes (including 79 COCO classes). We use the LVIS v1 validation set for evaluation and report performance on the base classes, novel classes, all classes, and the frequent/common/rare class splits defined by LVIS (i.e., frequent classes appear in 100 or more images, common classes appear in 10 to 100 images, and rare classes appear in 10 or fewer images). Note that all base classes are considered frequent classes.

\myparagraph{Comparisons}~Our main comparisons are with the recent state-of-the-art RNCDL method~\citep{Fomenko_2022_NeurIPS} due to its strong performance on large-scale benchmarks and its publicly available code. RNCDL uses a supervised loss and a SwAV loss on pseudo-labels during discovery. We compare with four versions of RNCDL, depending on if it uses a uniform prior for the generated pseudo-labels or a long-tailed prior, and if it uses the mask head of Mask R-CNN, or a simple version closer to Faster R-CNN (i.e., without masks). 
This makes RNCDL more comparable to PANDAS, which does not require masks or long-tailed priors.

\myparagraph{Evaluation}~To quantitatively evaluate NCD, we need to assign semantic class labels to the discovered novel classes. We follow related works on NCD~\citep{Fomenko_2022_NeurIPS,han2019learning,Han2020Automatically,fini2021unified,weng2021unsupervised} and use the Hungarian Algorithm~\citep{kuhn1955hungarian} for the matching procedure, which is described in Sec.~\ref{subsec:semantic-classes}. 
We then report the mean Average Precision (mAP) at an Intersection over Union (IoU) threshold of 0.5 for all experiments, unless stated otherwise. Reporting at an IoU of 0.5 is standard for VOC, and has been the recent standard for evaluating novel class generalization performance on LVIS~\citep{zhou2022detecting}. We report results on LVIS with IoU thresholds from 0.5 to 0.95 in an additional experiment in Sec.~\ref{subsec:results-coco-lvis}.

% define figure and table together to reduce white space
\begin{table}[t]
\begin{minipage}{0.49\linewidth}
    \begin{center}
      \caption{\textbf{Analyzing components of PANDAS on VOC.} Results of several flavors of PANDAS at mAP 0.5 with 250 clusters. Each result is the average over three runs. $^{\dagger}$ indicates an oracle version of our model. The default configuration of PANDAS uses a background classifier, a (Euclidean Dist.)$^{-2}$ similarity metric with $L_{1}$ probability normalization, Ground Truth (GT) prototypes for base classes, and cluster centers for novel classes. We include the difference of each variant with PANDAS (Default).
\label{tab:voc-add-studies}}
\centering
\resizebox{\linewidth}{!}{
\begin{tabular}{lccc}
\toprule
\textsc{PANDAS Variant} & mAP$_{base}$ & mAP$_{novel}$ & mAP$_{all}$ \\ 
\midrule
\rowcolor{Gray}
\multicolumn{4}{l}{\textit{Base and Novel Prototypes}}\\
\midrule
All Clusters & 57.9 \footnotesize{({\color{Red}-0.9})} & 20.5 \footnotesize{({\color{Red}-0.8})} & 39.2 \footnotesize{({\color{Red}-0.8})} \\
\textcolor{gray}{All GT Prototypes$^{\dagger}$} & \textcolor{gray}{58.0} \footnotesize{({\color{Red}-0.8})} & \textcolor{gray}{21.2} \footnotesize{({\color{Red}-0.1})} & \textcolor{gray}{39.6} \footnotesize{({\color{Red}-0.4})} \\
\midrule
\rowcolor{Gray}
\multicolumn{4}{l}{\textit{Similarity Metric \& Probability Normalization}}\\
\midrule
(Euclidean Dist.)$^{-2}$, Softmax & 46.6 \footnotesize{({\color{Red}-12.2})} & 17.1 \footnotesize{({\color{Red}-4.2})} & 31.8 \footnotesize{({\color{Red}-8.2})} \\
(Euclidean Dist.)$^{-1}$, $L_{1}$ & 57.4 \footnotesize{({\color{Red}-1.4})} & 18.9 \footnotesize{({\color{Red}-2.4})} & 38.2 \footnotesize{({\color{Red}-1.8})} \\
(Euclidean Dist.)$^{-1}$, Softmax & 56.6 \footnotesize{({\color{Red}-2.2})} & 20.5 \footnotesize{({\color{Red}-0.8})} & 38.6 \footnotesize{({\color{Red}-1.4})} \\
Dot Product Sim., $L_{1}$ & 57.3 \footnotesize{({\color{Red}-1.5})} & 13.9 \footnotesize{({\color{Red}-7.4})} & 35.6 \footnotesize{({\color{Red}-4.4})} \\
Dot Product Sim., Softmax & 60.3 \footnotesize{({\color{Green}+1.5})} & 13.7 \footnotesize{({\color{Red}-7.6})} & 37.0 \footnotesize{({\color{Red}-3.0})} \\
Cosine Sim., $L_{1}$ & 23.9 \footnotesize{({\color{Red}-34.9})} & 0.3 \footnotesize{({\color{Red}-21.0})} & 12.1 \footnotesize{({\color{Red}-27.9})} \\
Cosine Sim., Softmax & \textbf{60.4} \footnotesize{({\color{Green}+1.6})} & 15.0 \footnotesize{({\color{Red}-6.3})} & 37.7 \footnotesize{({\color{Red}-2.3})} \\
\midrule
\rowcolor{Gray}
\multicolumn{4}{l}{\textit{Use of a Background Classifier}}\\
\midrule
No Background Classifier & 50.6 \footnotesize{({\color{Red}-8.2})} & 11.8 \footnotesize{({\color{Red}-9.5})} & 31.2 \footnotesize{({\color{Red}-8.8})} \\
\midrule
\textbf{PANDAS (Default)} & 58.8 & \textbf{21.3} & \textbf{40.0} \\
\bottomrule
\end{tabular}
}

    \end{center}
	\end{minipage}\hfill 
	\begin{minipage}{0.49\linewidth}
    \begin{center}
      \caption{\textbf{Comparison to the state of the art on \COCOtoLVIS.} We report mAP at IoU of 0.5:0.95 for base, novel, and all classes. For PANDAS, we report the mean and standard deviation over three runs. For RNCDL~\citep{Fomenko_2022_NeurIPS}, we report a variant with a uniform prior and the standard method with a long-tailed prior. We provide a fully supervised oracle for comparison. $^*$ indicates our runs.
\label{tab:lvis-sota}}
\centering
\resizebox{\linewidth}{!}{
\begin{tabular}{lccc}
\toprule
\textsc{Method} & mAP$_{base}$ & mAP$_{novel}$ & mAP$_{all}$ \\ 
\midrule
\rowcolor{Gray}
\multicolumn{4}{l}{\textit{Image classification methods extended to detection}}\\
\midrule
Weng et al.~\citep{weng2021unsupervised} & 17.9 & 0.3 & 1.6 \\
ORCA~\citep{cao2022openworld} & 20.6 & 0.5 & 2.0 \\
UNO~\citep{fini2021unified} & \textbf{21.1} & \textbf{0.6} & \textbf{2.2} \\
\midrule
\rowcolor{Gray}
\multicolumn{4}{l}{\textit{Methods using boxes (i.e., Faster R-CNN)}}\\
\midrule
$k$-means & 17.8 & 0.2 & 1.6 \\
RNCDL (Uniform)$^*$ & 22.3 & 3.7 & 5.1 \\
RNCDL$^*$ & 23.1 & 4.2 & 5.6 \\
PANDAS (All Clusters) & 26.9\footnotesize{{\color{DarkGray}$\pm$0.1}} & 4.8\footnotesize{{\color{DarkGray}$\pm$0.2}} & 6.5\footnotesize{{\color{DarkGray}$\pm$0.2}} \\
PANDAS (No Background Cls.) & 27.9\footnotesize{{\color{DarkGray}$\pm$0.1}} & 5.0\footnotesize{{\color{DarkGray}$\pm$0.2}} & 6.7\footnotesize{{\color{DarkGray}$\pm$0.2}} \\
PANDAS (Ours) & \textbf{28.5}\footnotesize{{\color{DarkGray}$\pm$0.1}} & \textbf{5.1}\footnotesize{{\color{DarkGray}$\pm$0.3}} & \textbf{6.9}\footnotesize{{\color{DarkGray}$\pm$0.2}} \\
\midrule
\rowcolor{Gray}
\multicolumn{4}{l}{\textit{Methods using boxes and segmentation masks (i.e., Mask R-CNN)}}\\
\midrule
RNCDL (Uniform)$^*$ & 24.1 & 4.2 & 5.8 \\
RNCDL$^*$ & 23.1 & 4.9	& 6.3 \\
RNCDL & 25.0 & \textbf{5.4} & 6.9 \\
PANDAS (Ours) & \textbf{28.6}\footnotesize{{\color{DarkGray}$\pm$0.4}} & 5.2\footnotesize{{\color{DarkGray}$\pm$0.3}} & \textbf{7.0}\footnotesize{{\color{DarkGray}$\pm$0.3}} \\
\midrule
\textcolor{gray}{Fully Supervised (Oracle)} & \textcolor{gray}{39.4} & \textcolor{gray}{16.7} & \textcolor{gray}{18.5} \\
\bottomrule
\end{tabular}
}
    \end{center}
	\end{minipage}
\end{table}

% ----------------------------------------------------

\myparagraph{Implementation details}~To compute novel prototypes, PANDAS uses $k$-means clustering (implemented in FAISS~\citep{johnson2019billion}), which was chosen due to its scalability. The default version of PANDAS uses a similarity metric that is computed by inverting and squaring the squared-Euclidean distance from a feature to a prototype. We explore additional similarity metrics to validate this choice. Moreover, we divide the final classification logits of PANDAS by their $L_{1}$ norm to make them sum to one. We compare this to standard softmax normalization. Following \citet{Fomenko_2022_NeurIPS}, we initialize the R-CNN backbone with a ResNet-50 pre-trained on ImageNet-1k using the self-supervised MoCo v2 method~\citep{he2020momentum}. During base training, we update the final three backbone layers and all detection parameters of the R-CNN to maintain these generalized representations. Our main experiments with PANDAS use Faster R-CNN, which does not require mask annotations. However, we also report the results of PANDAS with mask annotations (i.e., using Mask R-CNN) on the \COCOtoLVIS benchmark to compare with RNCDL directly. For experiments with Mask R-CNN, we follow \citet{Fomenko_2022_NeurIPS} and use a class-agnostic mask head during base training and discovery. Additional details are in Sec.~\ref{supp:model-config}. 

% ----------------------------------------------------
\section{Results}

% ----------------------------------------------------
\subsection{Results on VOC}
\label{subsec:results-voc}

We use the smaller VOC dataset to validate our main design choices. First, we study the impact of the number of clusters used in the discovery phase, a hyper-parameter that can potentially have a crucial impact on the quality of NCD methods. We additionally assess the compute time used to run these experiments. We then study the impact of several design choices on the performance of PANDAS including the use of a background classifier, the choice of similarity metric, the probability normalization function, and the method used to compute base and novel prototypes. Finally, we provide qualitative outputs of PANDAS.

\myparagraph{Impact of the number of clusters}~The number of clusters used during the discovery phase can impact NCD model performance. To study this impact, we show the results of PANDAS and RNCDL using 10, 20, 50, 100, 250, 500, or 1000 novel clusters on VOC in Fig.~\ref{fig:num-clusters-voc} (numeric values are in Appendix Table~\ref{tab:voc-map-rncdl-pandas}). Note that VOC contains 20 total classes, with 10 base classes and 10 novel classes. For RNCDL, we found that using a uniform label prior yielded better results, so we report results using a uniform prior here and include results with a long-tailed prior in Appendix Table~\ref{tab:voc-map-rncdl-pandas}.
We observe that for both PANDAS and RNCDL, the performance on base classes remains relatively constant for various numbers of clusters. Performance on novel classes improves as the number of clusters grows, until performance plateaus. Overall, PANDAS outperforms RNCDL on novel classes, remains competitive on base classes, and has more stable results across runs (i.e., smaller standard deviations).

\myparagraph{Compute analysis}~To run the experiments in Fig.~\ref{fig:num-clusters-voc}, PANDAS required an average of 17 GPU minutes for 10 novel clusters and 210 GPU minutes for 1000 novel clusters (see Appendix Table~\ref{tab:time-voc} for timings). We ran all experiments on Tesla V100-SXM2-32GB GPUs for a fair comparison. In contrast, we found RNCDL required an average of 1463 GPU minutes, regardless of the number of clusters used. This indicates that, PANDAS not only outperforms RNCDL across various numbers of novel clusters, but also requires much less time. For example, when using 250 clusters, PANDAS outperforms RNCDL by 12\% mAP on novel classes while running over 36 times faster. The simplicity and speed of PANDAS make it easier to use in practice. For the remaining VOC experiments, PANDAS uses 250 novel clusters, which is a reasonable trade-off between accuracy and speed (i.e., more clusters lead to a more costly method). 

\begin{figure}[t!]
    \centering
        \includegraphics[height=0.22\linewidth]{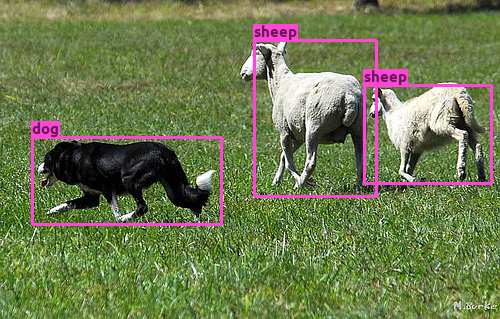}
        \includegraphics[height=0.22\linewidth]{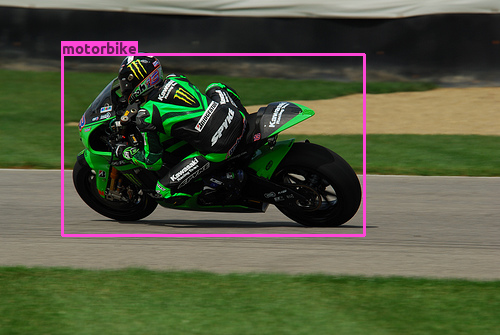}
        \includegraphics[height=0.22\linewidth]{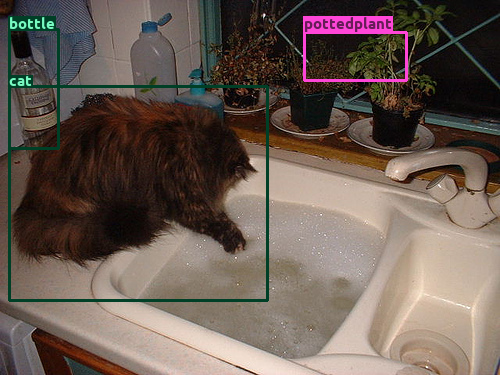}
%    \\
    \caption{\textbf{Qualitative results on VOC.} 
    Detections output by PANDAS. \basecolor{Base} classes include ``bottle'' and ``cat''. \novelcolor{Novel} classes include ``dog'', ``motorbike'', ``pottedplant'', and ``sheep''. Additional examples on VOC are in Sec.~\ref{supp:timing}.
    }
    \label{fig:voc-qual}
\end{figure}

\myparagraph{Validating the components of PANDAS}~Table~\ref{tab:voc-add-studies} shows the impact of the different choices we made for PANDAS regarding i) how to represent base classes, ii) how to represent novel classes, iii) the similarity metric used, iv) the probability normalization strategy (i.e., for converting similarity scores to probabilities between zero and one), and v) the use of a background classifier. Specifically, we compare the use of a background classifier to not using a background classifier. The background classifier is helpful in assigning regions discovered by PANDAS as background regions. Without the background classifier, PANDAS always assigns a box proposal to a discovered cluster. We then compare four similarity metrics ($\mathcal{S}$): an inverse Euclidean distance-squared metric, a squared inverse Euclidean distance-squared metric (the one used by the default PANDAS method), cosine similarity, and a dot product similarity computed as: $\left(\textbf{p} \cdot \textbf{f}\right)$ - 0.5$\left(\textbf{p} \cdot \textbf{p}\right)$, where $\textbf{p}$ is a prototype we are comparing to a feature $\textbf{f}$ and $\cdot$ is the dot product. Studying these similarity metrics and associated probability normalization schemes is crucial as PANDAS uses a distance-based classifier. Finally, our default PANDAS method uses ground truth prototypes for base classes and $k$-means clustering for discovering novel classes. We compare this to a variant that uses ground truth prototypes for base and novel classes (an oracle with access to novel class labels) and a variant using $k$-means clustering for all classes.

We find that the background classifier has a large impact on performance. We also find the similarity metric and associated probability normalization metric have large impacts. While cosine similarity and the dot product-based similarity work well for base classes, their performance on novel classes is much worse than our main PANDAS method that uses the inverse Euclidean distance-squared metric squared. While the variant using ground truth prototypes for base and novel classes has access to additional labels for novel classes, its performance is roughly the same as the variant that uses $k$-means clustering for both base and novel classes, as well as our main method. However, in Appendix Fig.~\ref{fig:pandas-vs-kmeans} we find that for fewer novel clusters, our main PANDAS method outperforms the variant using only clusters by large margins, especially on novel classes. This indicates that the all clusters variant requires more clusters to perform well. Overall, we see that all of our design choices are important and yield a method that exhibits strong performance.

\myparagraph{Visualizations from PANDAS}~Fig.~\ref{fig:voc-qual} shows detections from PANDAS. PANDAS remembers base classes (e.g., bottle, cat), while correctly detecting novel classes (e.g., dog, sheep, potted plant, motorbike). However, it struggles to detect multiple potted plants next to one another.

% ----------------------------------------------------
\subsection{Results on COCO to LVIS}
\label{subsec:results-coco-lvis}

\begin{table*}[t]
\caption{\textbf{Comparison with RNCDL on \COCOtoLVIS.} Performance is reported as mAP at IoU of 0.5 for the base, novel, all, frequent (f), common (c), and rare (r) classes. For RNCDL~\citep{Fomenko_2022_NeurIPS}, we report a variant with a uniform prior, the standard method with a long-tailed prior, a variant with Mask R-CNN, and a variant without Mask R-CNN. For PANDAS, we report a variant using only clusters, a variant without the background classifier, as well as the default method. For PANDAS, we report the mean and standard deviation over three runs.
\label{tab:main-results}}
\centering
\small
\begin{tabular}{lcccccc}
\toprule
\textsc{Method} & ~~mAP$_{base}$~~ & mAP$_{novel}$ & mAP$_{all}$ & ~~~mAP$_{f}$~~~ & ~~~mAP$_{c}$~~~ & ~~~mAP$_{r}$~~~ \\
\midrule
\rowcolor{Gray}
\multicolumn{7}{l}{\textit{Methods trained with bounding boxes (i.e., based on Faster R-CNN)}}\\
\midrule
RNCDL (Uniform) & 39.4 & 5.5 & 8.1 & 11.4 & 6.2 & 5.2 \\
RNCDL & 39.9 & 6.2 & 8.9 & 11.7 & 7.1 & 6.5 \\
PANDAS (All Clusters) & 42.5\footnotesize{{\color{DarkGray}$\pm$0.2}} & 8.4\footnotesize{{\color{DarkGray}$\pm$0.3}} & 11.0\footnotesize{{\color{DarkGray}$\pm$0.3}} & 14.2\footnotesize{{\color{DarkGray}$\pm$0.1}} & 8.6\footnotesize{{\color{DarkGray}$\pm$0.3}} & 10.1\footnotesize{{\color{DarkGray}$\pm$1.2}} \\
PANDAS (No Background Cls.) & 44.5\footnotesize{{\color{DarkGray}$\pm$0.3}} & 8.5\footnotesize{{\color{DarkGray}$\pm$0.3}} & 11.3\footnotesize{{\color{DarkGray}$\pm$0.3}} & 14.3\footnotesize{{\color{DarkGray}$\pm$0.1}} & 8.9\footnotesize{{\color{DarkGray}$\pm$0.5}} & 10.3\footnotesize{{\color{DarkGray}$\pm$1.0}} \\
PANDAS (Ours) & \textbf{45.6}\footnotesize{{\color{DarkGray}$\pm$0.2}} & \textbf{8.7}\footnotesize{{\color{DarkGray}$\pm$0.5}} & \textbf{11.5}\footnotesize{{\color{DarkGray}$\pm$0.4}} & \textbf{14.7}\footnotesize{{\color{DarkGray}$\pm$0.1}} & \textbf{9.2}\footnotesize{{\color{DarkGray}$\pm$0.6}} & \textbf{10.4}\footnotesize{{\color{DarkGray}$\pm$1.5}} \\
\midrule
\rowcolor{Gray}
\multicolumn{7}{l}{\textit{Methods trained with bounding boxes and segmentation masks (i.e., based on Mask R-CNN)}}\\
\midrule
RNCDL (Uniform) & 40.7 & 5.6 & 8.3 & 11.9 & 5.9 & 6.3 \\
RNCDL & 39.6 & 7.4 & 9.9 & 11.8 & 8.0 & 10.3 \\
PANDAS (Ours) & \textbf{45.4}\footnotesize{{\color{DarkGray}$\pm$0.6}} & \textbf{8.9}\footnotesize{{\color{DarkGray}$\pm$0.5}} & \textbf{11.7}\footnotesize{{\color{DarkGray}$\pm$0.5}} & \textbf{14.9}\footnotesize{{\color{DarkGray}$\pm$0.2}} & \textbf{8.9}\footnotesize{{\color{DarkGray}$\pm$0.2}} & \textbf{11.3}\footnotesize{{\color{DarkGray}$\pm$1.7}} \\
\bottomrule
\end{tabular}
\end{table*}

Next, we experimentally validate PANDAS on the more realistic \COCOtoLVIS benchmark. For RNCDL, we use 3000 clusters for novel classes since this worked best in \citet{Fomenko_2022_NeurIPS} and run experiments once due to long run-times. For PANDAS, we use 5000 novel clusters, which provides a nice trade-off between performance and speed. We run PANDAS three times with different seeds and report the mean and standard deviation over runs. We provide results for additional numbers of novel clusters for PANDAS and RNCDL in Appendix Table~\ref{tab:lvis-num-clusters-0.5}.

We report our main results on \COCOtoLVIS in Table~\ref{tab:main-results}, where we compare RNCDL with a uniform prior and a long-tail prior, as well as with mask annotations and without. We also compare four variants of PANDAS: one using only clusters without base prototypes, one that does not use the background classifier, our default PANDAS method with Faster R-CNN (without mask annotations), and our default PANDAS method with Mask R-CNN (with mask annotations). Overall, we find that PANDAS consistently outperforms RNCDL across all class groups. This is remarkable as the main variant of RNCDL assumes a long-tail class prior, while PANDAS make no assumptions about the label distribution. Moreover, PANDAS does not require the inclusion of additional segmentation mask annotations, which provide large performance benefits for RNCDL, especially on rare classes (i.e., masks provide a 3.8\% mAP gain on rare classes). We find that the performance of PANDAS degrades slightly when using only clusters (i.e., 0.3\% mAP drop on novel classes) or when the background classifier is not used (i.e., 0.2\% mAP drop on novel classes). When using mask annotations, the performance of PANDAS remains roughly the same, with a slight improvement on rare classes (i.e., 0.9\% mAP gain). Moreover, PANDAS is relatively stable across runs, with the smallest standard deviation on frequent classes of 0.1\%, and the largest on rare classes of 1.5\%. Finally, while RNCDL required roughly 4296 GPU minutes on LVIS, PANDAS required an average of just 806 minutes, which is a fivefold speed improvement.

\myparagraph{Comparison with the state of the art}~In addition to our comparisons with RNCDL, we also compare with state-of-the-art NCD methods from classification that were extended to detection in \citet{Fomenko_2022_NeurIPS}. These include:

\begin{itemize}[noitemsep, nolistsep]
   \item $\bm{k}$\textbf{-means}, a baseline from \citet{Fomenko_2022_NeurIPS} that clusters RPN-predicted boxes using $k$-means. Test boxes are given the label of the closest cluster with 100\% confidence.
   \item \textbf{Weng et al.}~\citep{weng2021unsupervised}, which learns representations using contrastive learning and then applies $k$-means clustering.
   \item \textbf{ORCA}~\citep{cao2022openworld}, which uses a mechanism to reduce bias towards known classes, pseudo-labeling with pairwise similarities, and regularization to avoid cluster collapse.
   \item \textbf{UNO}~\citep{fini2021unified}, which learns using a unified cross-entropy objective function. It uses ground truth labels, as well as pseudo-labels for unlabeled examples from SwAV~\citep{caron2020unsupervised}.
\end{itemize}

% define figure and table together to reduce white space
\begin{table}[t]
\begin{minipage}{0.49\linewidth}
    \begin{center}
          \centering
    \includegraphics[trim=5 10 5 5, clip, width=\linewidth]{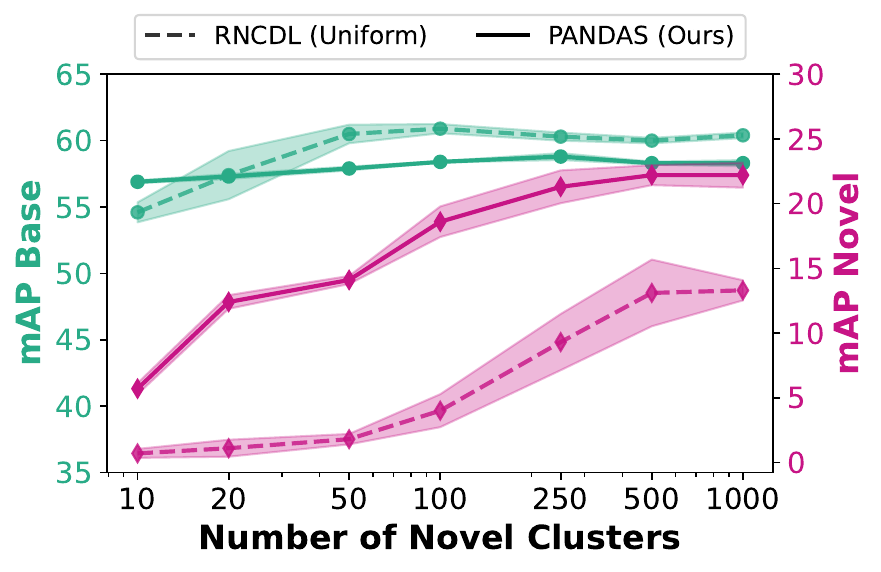}
    \captionof{figure}{\textbf{NCD Performance vs. Clusters on VOC.} We study NCD performance as a function of the number of novel clusters. Each curve is averaged over three runs with shaded standard deviations.
    }
    \label{fig:num-clusters-voc}
    \end{center}
	\end{minipage}\hfill
	\begin{minipage}{0.49\linewidth}
    \begin{center}
              \centering
        \includegraphics[trim=20 -80 15 -95, clip, width=\linewidth]{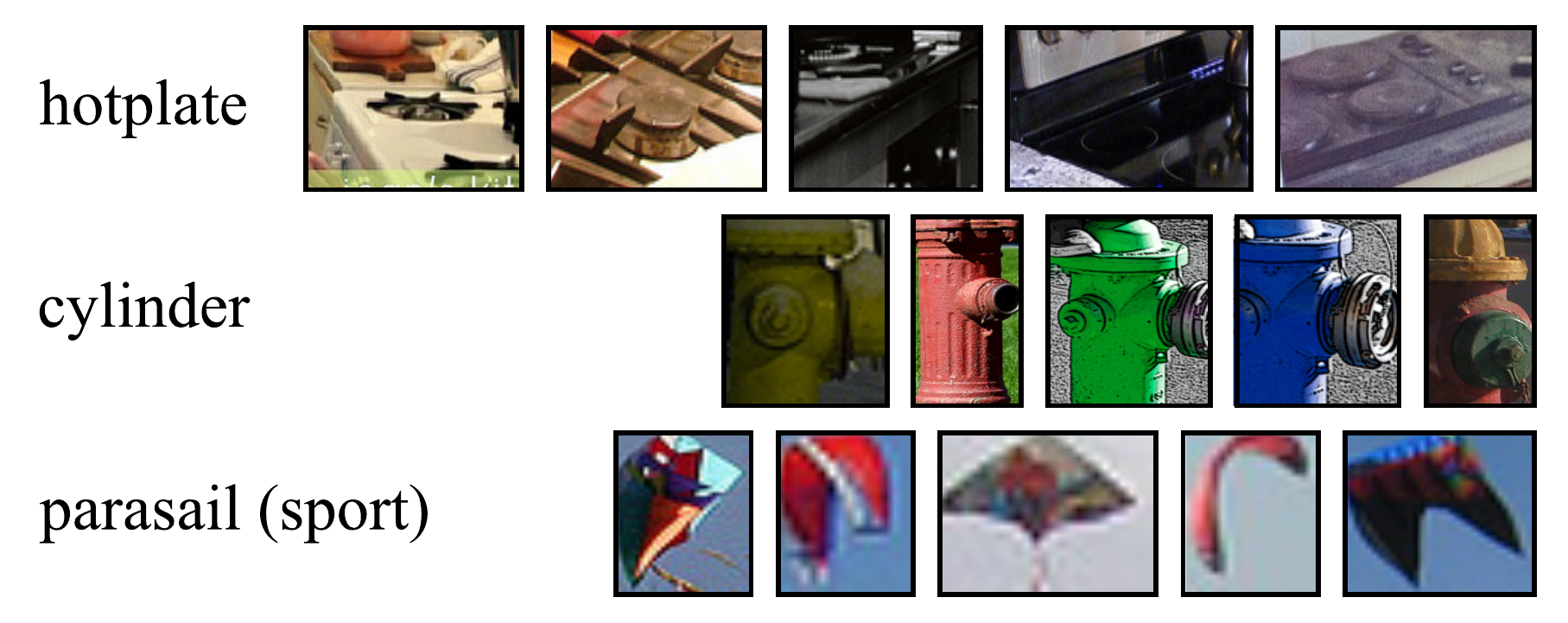} 
    \captionof{figure}{
    \textbf{Clusters on \COCOtoLVIS.} For three discovered clusters, we show the five boxes closest to cluster centers and indicate the class name they were mapped to during assignment.
    }
    \label{fig:lvis-cluster-crops}
    \end{center}
	\end{minipage}
\end{table}

We compare with these state-of-the-art methods in Table~\ref{tab:lvis-sota} using mAP at IoU of 0.5 to 0.95, as reported in \citet{Fomenko_2022_NeurIPS}. Overall, we see that NCD classification methods that have been extended to detection perform quite poorly, especially on novel classes, where none of the methods achieve 1\% mAP. PANDAS outperforms RNCDL with both a uniform and a long-tailed prior when only box annotations are used by 0.9\% mAP and 1.3\% mAP on novel and all classes, respectively. When RNCDL uses mask annotations, PANDAS performs comparably to it regardless of if PANDAS uses mask annotations or not. However, obtaining mask annotations for additional datasets is costly, and it is not always an option in practical scenarios.

\myparagraph{Visualizations from PANDAS}~Since PANDAS discovers clusters, we can visualize the closest training examples to different cluster centers to better understand what has been learned by the model. In Fig.~\ref{fig:lvis-cluster-crops}, we show the five closest training examples (box crops) to three cluster centers which were mapped to the ``hotplate'', ``cylinder'', and ``parasail (sports)'' classes using Hungarian Matching. We see that there is some confusion between hotplates and stoves, which is not surprising as the two classes are semantically similar and PANDAS was trained in an unsupervised way.

\begin{figure}[t!]
        \centering
        \includegraphics[height=0.22\linewidth]{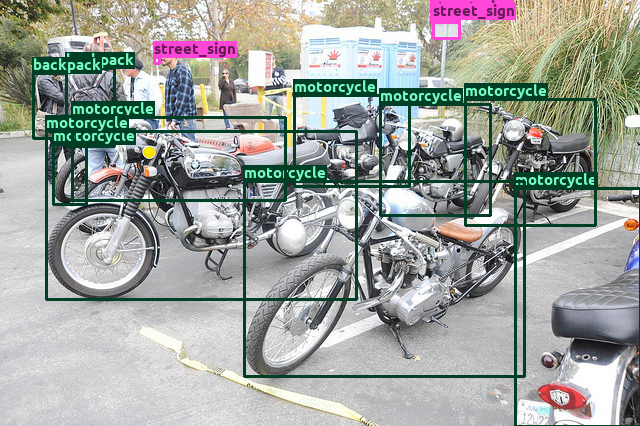}
        \includegraphics[height=0.22\linewidth]{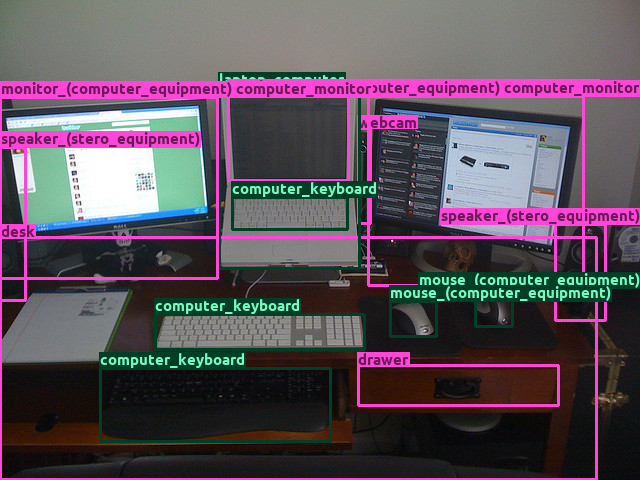}
        \includegraphics[height=0.22\linewidth]{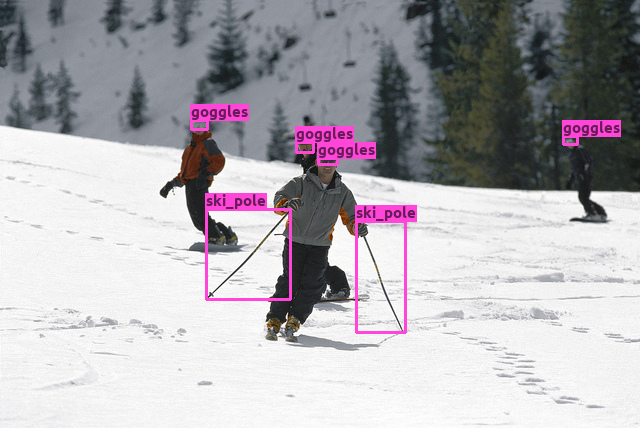}
        \\
        \includegraphics[height=0.22\linewidth]{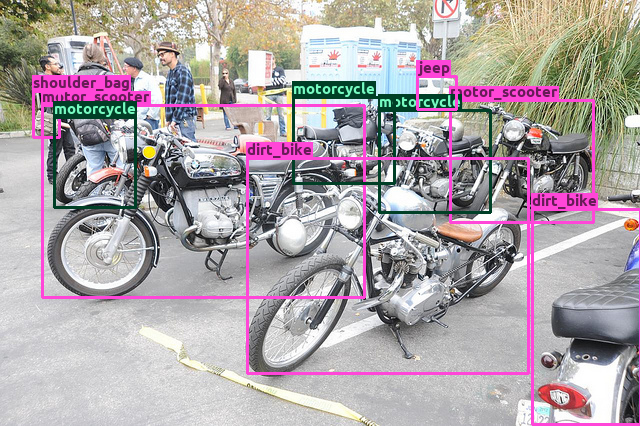}
        \includegraphics[height=0.22\linewidth]{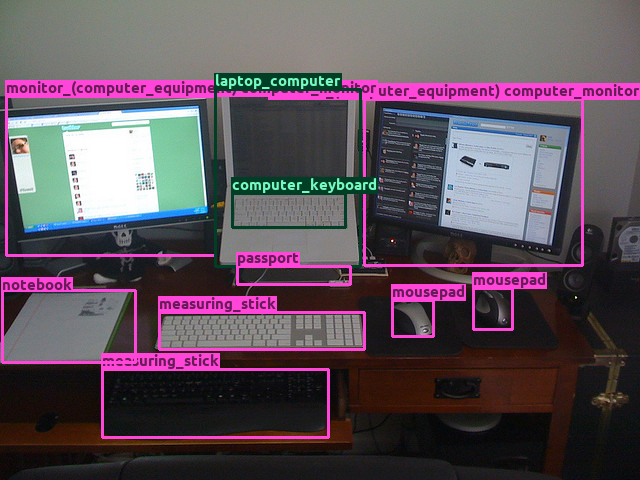}
        \includegraphics[height=0.22\linewidth]{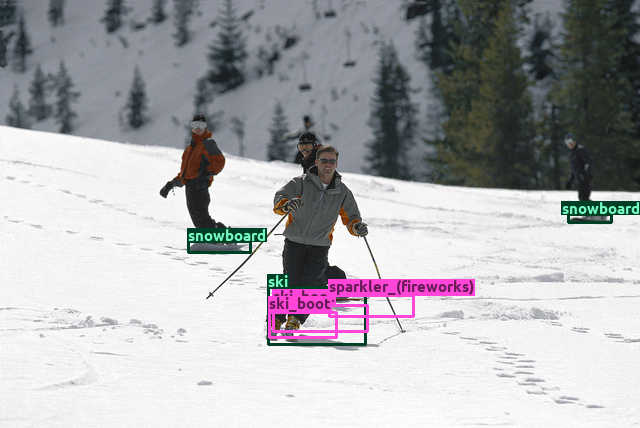}
    \caption{\textbf{Qualitative results on \COCOtoLVIS.} Ground truth (top) and PANDAS outputs (bottom) with \basecolor{Base} and \novelcolor{Novel} classes. Additional examples on \COCOtoLVIS are in Sec.~\ref{supp:results-map-0.5-0.95}.
    }
    \label{fig:lvis-qual}
\end{figure}

In Fig.~\ref{fig:lvis-qual} we show ground truth detections, as well as detections predicted by PANDAS on three images from LVIS. Visualizing these predictions can help us better understand PANDAS. For example, in the computer image, we see that it correctly detects the laptop (\basecolor{base}), computer keyboard (\basecolor{base}), and computer monitors (\novelcolor{novel}), but it confuses the computer mouse (\basecolor{base}) with a mousepad (\novelcolor{novel}). Moreover, in the motorcycle image, we can see the difficulty of LVIS, requiring models to distinguish multiple fine-grained object categories (e.g., motorcycle vs. dirtbike, vs. motorscooter). These semantically similar objects prove difficult for our unsupervised method. Interestingly, we find that in several cases, PANDAS discovers objects not included in the ground truth. For example, it discovers a notebook (\novelcolor{novel}) in the computer image, and skis (\basecolor{base}), ski boots (\novelcolor{novel}), and snowboards (\basecolor{base}) in the skiing image. This is interesting, as NCD methods could be used in the future to automatically provide annotations for images.

% ----------------------------------------------------
\section{Discussion \& Conclusion}
Novel class discovery and detection is a recently proposed paradigm that challenges models to learn from a labeled base dataset and then discover novel classes in an unsupervised way. In this paper, we introduced PANDAS, a prototype-based method that uses a distance-based classifier to detect novel objects. PANDAS performs favorably to recent state-of-the-art NCD methods from classification and detection, while running much faster and without requiring additional information (e.g., mask annotations, novel class label distribution, or number of novel classes). This makes PANDAS an ideal candidate for practical applications, where manual annotations are not readily available and the number or distribution of novel objects is unknown.

Moreover, compared to the recent RNCDL method, PANDAS does not require self-supervised SwAV training with pseudo-labels, a memory module, or storing base images to perform a supervised loss update during discovery. This makes PANDAS much simpler to implement and faster to use in practice (e.g., PANDAS was over 36 times faster on VOC and over 
five times faster on LVIS compared to RNCDL). We attribute the strong performance of PANDAS to its distance-based classifier, which operates on prototypes that lie in the same feature space. That is, PANDAS does not require separate base and novel classifiers. Finally, the background classifier of PANDAS yields significant performance improvements on smaller datasets like VOC.

Since PANDAS operates on feature vectors, it can easily be extended to any task where a feature representation is available, be it at the sample level (e.g., classification, detection) or pixel level (e.g., semantic segmentation). PANDAS can easily be applied in alternative two-stage detectors by extracting box features from the box proposals produced in the first stage, and then performing distance-based classification in place of the second stage. For transformer-based detectors like DETR~\citep{carion2020end}, one could potentially apply PANDAS on the query-conditioned features output by the transformer decoder. Naturally, the effectiveness in these scenarios would require a thorough analysis, but we are optimistic about the prospect of applying PANDAS in additional detection architectures and on additional problems (e.g., segmentation). Moreover, as with other NCD methods, PANDAS could be used more generally to annotate new objects in images, or as part of a pipeline to flag new objects of interest for inspection by annotators.

One interesting area for future research could include the development of NCD methods for detection that update their feature representations during the discovery phase. While PANDAS and RNCDL use frozen features after the base phase, updating features during the discovery phase could allow the network to learn representations that are more tailored to new classes. The challenge in updating representations is handling feature drift, especially when it is not possible to store the entire base dataset. That is, catastrophic forgetting of previously learned concepts is a risk that must be mitigated when designing new methods.

Another interesting future direction could include the use of a stronger base detector. Strong pre-trained models have become more popular in recent years~\citep{radford2021learning,caron2021emerging} and using them as a starting point for NCD is likely to improve performance even further. In this study, we focused on training the base detector on smaller datasets for two reasons: i) to maintain consistency with prior work~\citep{Fomenko_2022_NeurIPS} for a fair and direct comparison and ii) to avoid class leakage (i.e., when novel classes are seen during base pre-training). This allowed us to compare PANDAS with prior work and study its performance in an isolated NCD setting without leakage. In practical scenarios, it would be ideal to optimize performance by starting from a strong pre-trained model. If future studies use a strong pre-trained detector, we recommend using the same detector for all NCD models for a fair comparison.

\myparagraph{Acknowledgements} We thank Yannis Kalantidis, Grégory Rogez, and Mingxuan Liu for their comments and useful discussions throughout the project.

% ----------------------------------------------------
\bibliography{egbib}
\bibliographystyle{collas2024_conference}

% ----------------------------------------------------
\newpage
\appendix
\section{Appendix}
\beginsupplement

% ------------------------------------------------
\section{Implementation Details}
\label{supp:model-config}

PANDAS builds on the Faster R-CNN model implemented in the torchvision package from PyTorch. We use a ResNet-50 backbone~\citep{He_2016_CVPR} pre-trained on ImageNet-1k using MoCo v2~\citep{he2020momentum} and a Feature Pyramid Network (FPN)~\citep{lin2017feature}. We use the MoCo v2 weights trained using 800 epochs from the MoCo GitHub repository (\url{https://github.com/facebookresearch/moco}). These are the exact weights used by RNCDL~\citep{Fomenko_2022_NeurIPS}.

We replace the per-class regressor of Faster R-CNN with a class-agnostic regressor. During training, we only update the last 3 layers of the backbone and all detection-specific parameters of the Faster R-CNN.

For the supervised base phase on VOC, we train PANDAS using stochastic gradient descent (SGD) with momentum of 0.9 for 85 epochs with a learning rate of 0.02 and weight decay of 0.0001 on a single GPU. We perform a warmup phase for 350 iterations where we linearly increase the learning rate from 0.001 to 0.02. During training, we drop the learning rate by a factor of 10 at epochs 16 and 22. We use the default Faster R-CNN parameter settings from torchvision (for the COCO dataset) for all other parameters (\url{https://github.com/pytorch/vision/blob/main/references/detection/train.py}). We exclusively use horizontal flip data augmentations. The checkpoint trained on the base dataset of VOC using these parameters achieves an mAP of 30.5\% at an IoU 0.5 on the full VOC test set.

For the supervised base phase on COCO with Faster R-CNN, we train PANDAS on the \COCOhalf~training dataset defined by \citet{Fomenko_2022_NeurIPS}. We train using SGD with momentum of 0.9 for 26 epochs, with a learning rate of 0.02 and weight decay of 0.0001 on a single GPU. We perform a warmup phase for 1000 iterations where we linearly increase the learning rate from 0.00002 to 0.02. During training, we drop the learning rate by a factor of 10 at epochs 16 and 22. Similar to VOC, we use the default COCO detection parameters from torchvision for all other parameters and exclusively use horizontal flip data augmentations. The checkpoint trained on COCO using these parameters achieves an mAP of 51.8\% at an IoU of 0.5 on the \COCOhalf~validation dataset. For the supervised base phase on COCO with a class-agnostic variant of Mask R-CNN, we use the same parameter settings and augmentations with a batch size of 12. The checkpoint trained on COCO with these parameters achieves an mAP of 52.0\% at an IoU of 0.5 on the \COCOhalf~validation dataset.

For the discovery phase, we extract RPN boxes from the discovery dataset using our model parameters (i.e., checkpoint) trained on the base dataset. Specifically, we extract the top 100 boxes output by the network per image that have a score higher than 0.05, which is standard for COCO detection. We then use the $k$-means clustering implementation from FAISS~\citep{johnson2019billion} to cluster boxes. On VOC, we set the maximum number of clustering iterations to 1000 and the maximum number of retries to 10. For large-scale experiments on \COCOtoLVIS, we found that using 250 maximum iterations and 5 maximum retries performed comparably, while running in less time, so we use these parameters for LVIS. For evaluation on VOC, we use the standard COCO evaluation procedure that keeps the top 100 boxes per image and filters boxes with a score less than 0.05. For LVIS, we follow the standard procedure that keeps the top 300 boxes per image and do not filter any boxes based on scores~\citep{gupta2019lvis}.

\myparagraph{RNCDL implementation details}~For RNCDL, we use the publicly available implementation on GitHub (\url{https://github.com/vlfom/RNCDL}). For all experiments on VOC, we do not use mask annotations, in order to make RNCDL directly comparable to PANDAS. For the supervised base phase on VOC, we trained the model for 19720 iterations with a 350 iteration linear warmup phase (with a warmup factor of 0.001) and learning rate drops at iterations 3712 and 5104 (which is roughly equivalent to training for 85 epochs with learning rate drops at 16 and 22 epochs, which were the parameters used by PANDAS for the base phase). All other parameter values were kept the same as in \citet{Fomenko_2022_NeurIPS}. In experiments with 100 novel clusters, we found these modified base training parameters yielded higher novel class performance and similar base class performance for RNCDL than the default parameters reported in \citep{Fomenko_2022_NeurIPS}, so we used the modified parameters here. This base model achieved 30.8\% mAP at an IoU of 0.5 on the full VOC test set. For the discovery phase on VOC and all experiments on \COCOtoLVIS, we use the parameter settings from \citet{Fomenko_2022_NeurIPS}.

\myparagraph{Comparison with the state of the art}~For comparisons with 
state-of-the-art methods in Table~\ref{tab:lvis-sota}, we report our runs of each method with a $^{*}$. Besides PANDAS, for all other methods and runs, we report the performance values from \citet{Fomenko_2022_NeurIPS}.

% ------------------------------------------------
\section{VOC Dataset Distributions}

\begin{figure*}[t!]
    \centering
    \begin{subfigure}[t]{0.33\linewidth}
       \centering
        \includegraphics[width=\linewidth]{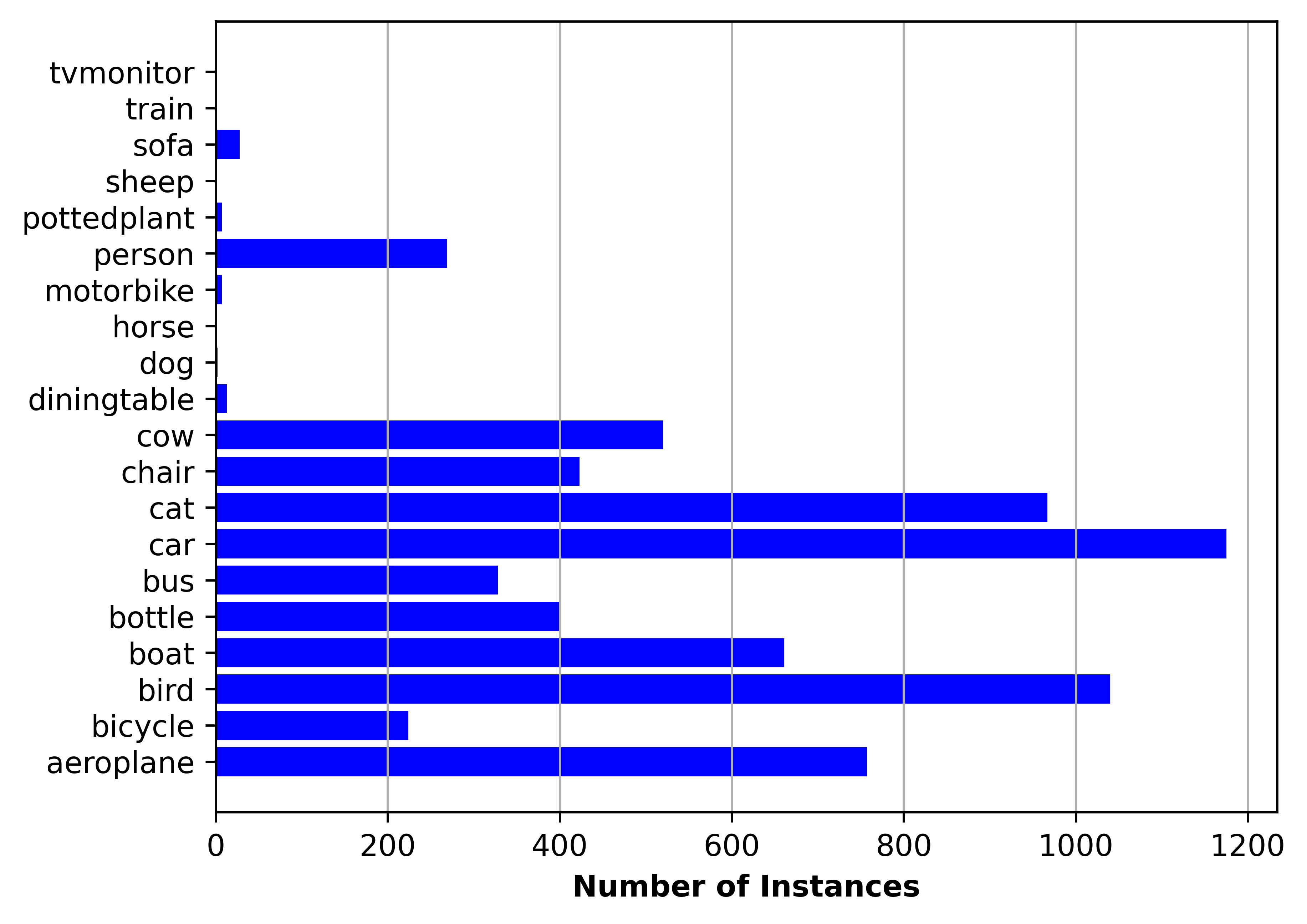}
        \caption{Base Dataset}
   \end{subfigure} 
    \begin{subfigure}[t]{0.33\linewidth}
        \centering
        \includegraphics[width=\linewidth]{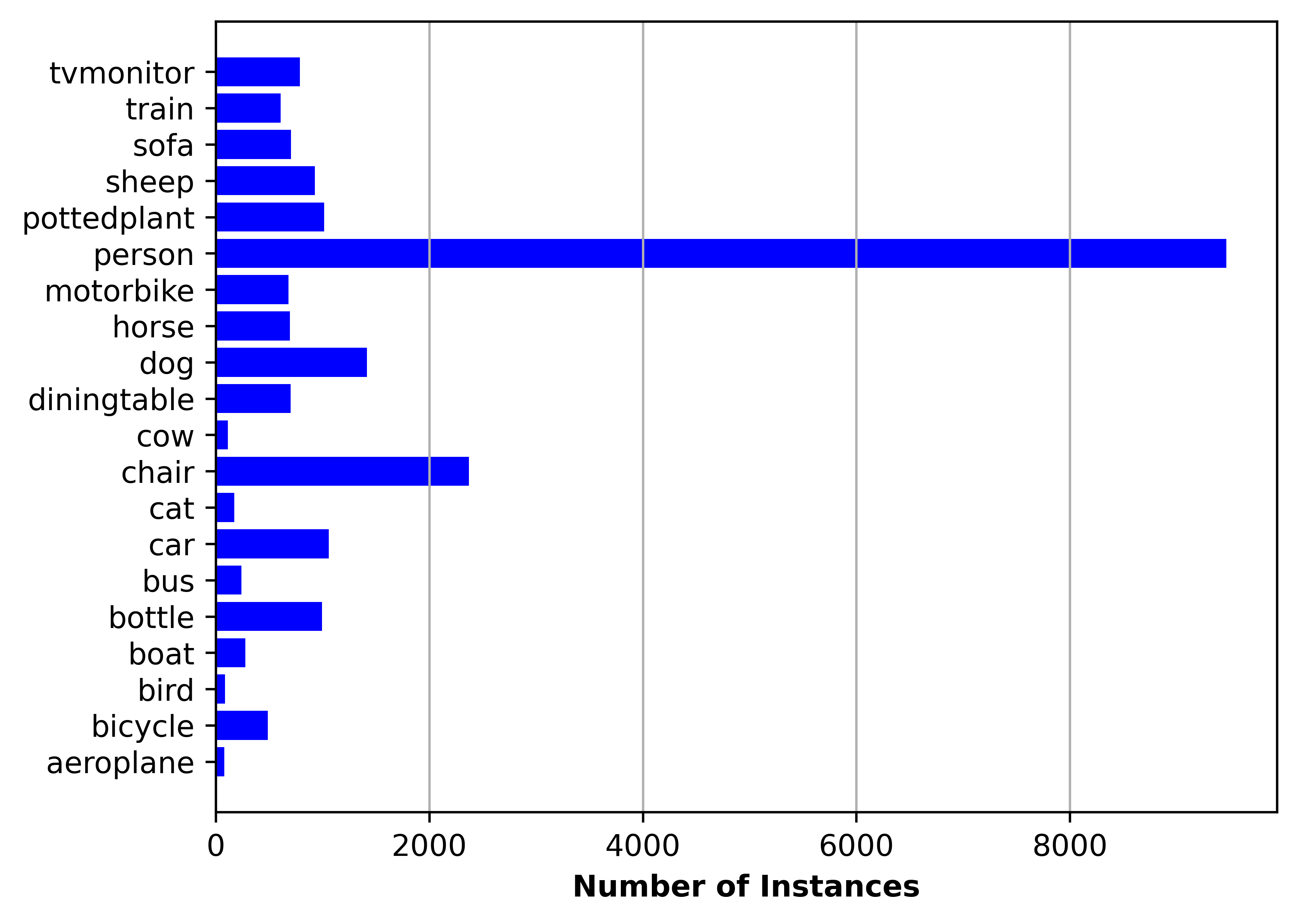}
        \caption{Discovery Dataset}
    \end{subfigure} 
    \begin{subfigure}[t]{0.33\linewidth}
       \centering
        \includegraphics[width=\linewidth]{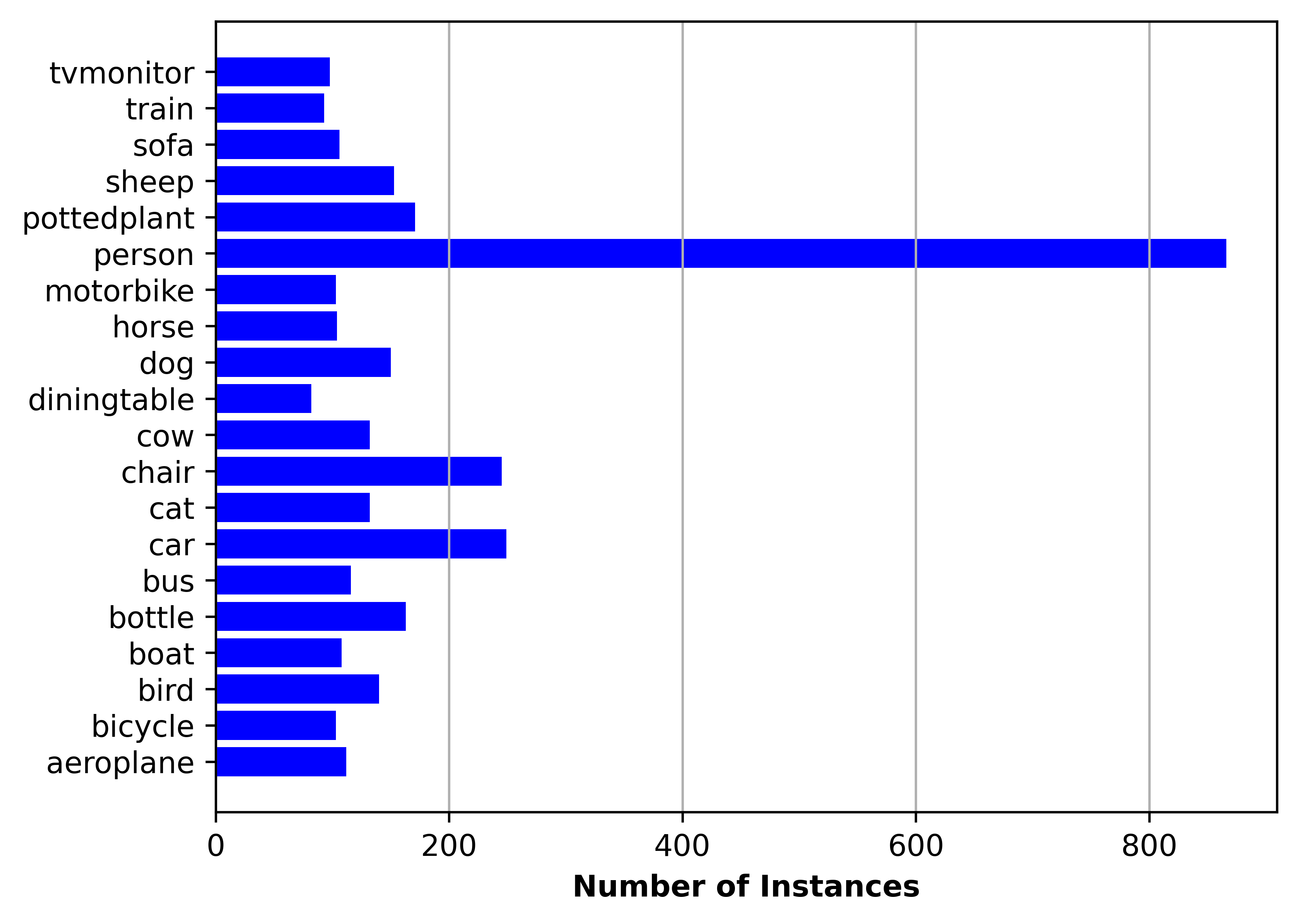}
        \caption{Test Dataset}
   \end{subfigure} 
   \\
    \caption{\textbf{Distributions of the VOC Dataset.} We report the number of object instances for each class included in the (a) Base Dataset, (b) Discovery Dataset, and (c) Test Dataset (composed of validation images). We withhold annotations for non-base classes included in the base dataset during the base phase.
    }
    \label{fig:voc-distributions}
\end{figure*}

In Fig.~\ref{fig:voc-distributions}, we show the distributions of class instances from the base dataset, the discovery dataset, and the test set for the VOC split that we use in experiments, which is the ``disjoint'' split originally from \citet{cermelli2022incremental}. Note that VOC contains 20 total classes with 3706 images in the base dataset, 6876 images in the discovery dataset, and 1449 images in the validation dataset (used for testing). In the base dataset, the following 10 classes are considered \basecolor{base}: \textit{aeroplane}, \textit{bicycle}, \textit{bird}, \textit{boat}, \textit{bottle}, \textit{bus}, \textit{car}, \textit{cat}, \textit{chair}, and \textit{cow}. Only ground truth annotations for object instances in this base class set are provided during the base phase (i.e., annotations for object instances from classes outside of this set are not provided during the base phase). The discovery dataset contains object instances from all 20 classes, where the following 10 classes are considered \novelcolor{novel}: \textit{diningtable}, \textit{dog}, \textit{horse}, \textit{motorbike}, \textit{person}, \textit{pottedplant}, \textit{sheep}, \textit{sofa}, \textit{train}, and \textit{tvmonitor}. The validation set of VOC is used as the test set during experiments, and contains object instances from all 20 classes.

% ------------------------------------------------
\section{Additional Results on VOC}
\label{supp:timing}

% ------------------------------------------------
\myparagraph{Efficiency analysis}~In Table~\ref{tab:time-voc}, we provide time estimates, in minutes, for the VOC results reported in Fig.~\ref{fig:num-clusters-voc} for RNCDL (Uniform) and PANDAS. However, the GitHub code for RNCDL does not run on fewer than two GPUs, and the default configuration uses four GPUs, which is the one we used for all experiments. Since PANDAS runs on a single GPU, we multiplied the time estimates for RNCDL by four to estimate the time required for RNCDL to run on a single GPU, which is comparable to PANDAS. We believe this is a reasonable way to estimate the single GPU run-time of RNCDL since running across four GPUs requires a forward pass (to compute the loss) and a backward pass (to compute gradients) per batch per GPU. Finally, note that the clustering stage of PANDAS is run on CPU and not GPU. All experiments were run on Tesla V100-SXM2-32GB GPUs for fair comparison.

Overall, RNCDL requires roughly the same amount of compute time, regardless of the number of novel clusters used (i.e., roughly 1463.3 minutes). In contrast, the compute time for PANDAS increases as the number of novel clusters increases, but is overall much less than RNCDL. For example, PANDAS is 87 times faster than RNCDL when using 10 novel clusters, and still seven times faster when using 1000 novel clusters. In Fig.~\ref{fig:num-clusters-voc}, we find that the performance of RNCDL and PANDAS plateaus around 1000 novel clusters, so we do not run experiments with larger numbers of novel clusters.

In addition to PANDAS' strong performance and lower compute time compared to RNCDL, PANDAS also does not require storing the base dataset during the discovery phase to prevent catastrophic forgetting of base classes. This is because PANDAS only stores prototypes for base classes and does not perform gradient descent updates, which are known to cause catastrophic forgetting of previous knowledge~\citep{mccloskey1989}. Since PANDAS does not store the base dataset, it is more memory efficient than RNCDL. Note that the base output layer and the novel output layer of RNCDL require the same amount of storage as the prototypes from PANDAS. This additional storage requirement could limit the applications where RNCDL could be used (e.g., data can be sensitive or private and not eligible for storage). Moreover, updating a model on previous (base) data during the discovery phase can add extra compute time, in comparison to only updating on discovery data.

% ------------------------------------------------
\myparagraph{Numeric values for experiments}~In Table~\ref{tab:voc-map-rncdl-pandas}, we provide numeric mAP values for the experiments from Fig.~\ref{fig:num-clusters-voc}, as well as mAP values for the long-tailed version of RNCDL. On base classes, the uniform variant of RNCDL (Uniform) outperforms the long-tail variant when fewer novel clusters are used, and the two variants have comparable performance when more novel clusters are used. On novel classes, the uniform variant of RNCDL consistently outperforms the long-tailed variant (with the exception of 10 novel clusters where performance is similar). This is likely because VOC is a more balanced dataset than LVIS. We find that the performance of PANDAS is competitive with RNCDL on base classes, while exhibiting large performance improvements on novel classes across various numbers of novel clusters. As an additional experiment, we evaluate the performance of PANDAS when using more clusters to determine the point at which its performance starts to degrade. We ran PANDAS with three seeds using 2500 and 5000 novel clusters, where novel class performance was 20.7\% mAP and 21.9\% mAP, respectively. These results can be directly compared to those in Table~\ref{tab:voc-map-rncdl-pandas} and indicate that the novel class performance of PANDAS starts to degrade when using more than 1000 clusters. However, note that using more clusters is not always desirable as it increases compute time.

\begin{figure}[t]
    \centering
     \includegraphics[width=0.5\linewidth]{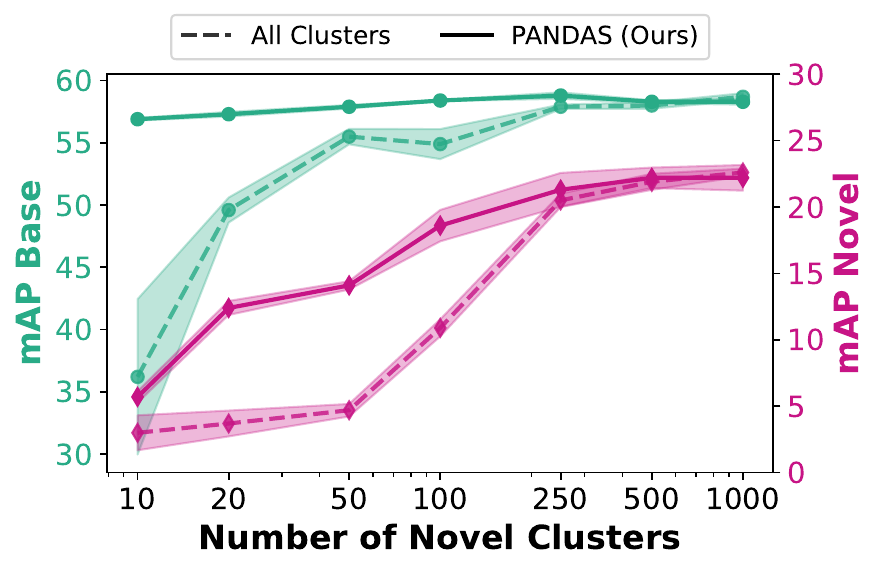}
     \caption{
     NCD performance as a function of the number of novel clusters on VOC for PANDAS and a variant of PANDAS that uses All Clusters (i.e., no ground truth base prototypes). Standard deviation is shown by the shaded regions.
     }
     \label{fig:pandas-vs-kmeans}
\end{figure}

% ------------------------------------------------
\myparagraph{PANDAS with all clusters}~In Fig.~\ref{fig:pandas-vs-kmeans}, we compare the performance of our default PANDAS method, which uses ground truth prototypes for base classes and cluster centers for novel classes, with a variant of PANDAS that uses cluster centers for all classes (All Clusters). On base classes, performance of our default PANDAS method remains relatively constant across various numbers of novel clusters. In contrast, the All Clusters version of PANDAS performs poorly on base classes if it does not have a large enough number of cluster centers. This evidence further supports our idea of using ground truth prototypes for base classes, i.e., prototypes computed using the annotations provided for the base classes in the training set. On novel classes, we see a similar trend where our default PANDAS method outperforms the All Clusters variant, unless more clusters are used. This indicates that the ground truth base prototypes help PANDAS on both base and novel classes, especially when fewer prototypes are used.

\label{supp:dataset-distributions}
\begin{table}[t]
\caption{Time estimates for VOC experiments (in minutes) averaged over three runs for various numbers of novel clusters. PANDAS was run on a single GPU and we report its time. RNCDL was run on four GPUs in parallel, so we report the time to run the experiments multiplied by four for comparison with PANDAS.
\label{tab:time-voc}}
\centering
\begin{tabular}{lccccccc}
\toprule
\textsc{Novel Clusters} & \textsc{10} & \textsc{20} & \textsc{50} & \textsc{100} & \textsc{250} & \textsc{500} & \textsc{1000} \\ 
\midrule
RNCDL (Uniform) & 1482.4 & 1544.4 & 1411.6 & 1425.6 & 1478.0 & 1412.4 & 1488.4 \\
PANDAS (Ours) & 17.0 & 20.6 & 21.8 & 21.5 & 40.5 & 110.4 & 209.7 \\
\bottomrule
\end{tabular}
\end{table}

\begin{table*}[t]
\caption{Numeric mAP values (corresponding to Fig.~\ref{fig:num-clusters-voc}) for experiments on VOC averaged over three runs across various numbers of novel clusters. We report results for RNCDL using a uniform prior (Uniform) and the standard long-tail prior.
\label{tab:voc-map-rncdl-pandas}}
\centering
\resizebox{\linewidth}{!}{
\begin{tabular}{lcccccccccccccc}
\toprule
& \multicolumn{7}{c}{mAP$_{base}$} & \multicolumn{7}{c}{mAP$_{novel}$} \\
\cmidrule(r){2-8} \cmidrule(r){9-15}
\textsc{Novel Clusters} & \textsc{10} & \textsc{20} & \textsc{50} & \textsc{100} & \textsc{250} & \textsc{500} & \textsc{1000} & \textsc{10} & \textsc{20} & \textsc{50} & \textsc{100} & \textsc{250} & \textsc{500} & \textsc{1000} \\ 
\midrule
RNCDL & 50.1 & 51.7 & 54.2 & 58.9 & \textbf{60.8} & \textbf{60.8} & \textbf{61.0} & 0.8 & 0.8 & 0.5 & 2.2 & 4.7 & 8.6 & 11.2 \\
RNCDL (Uniform) & 54.6 & \textbf{57.4} & \textbf{60.5} & \textbf{60.9} & 60.3 & 60.0 & 60.4 & 0.7 & 1.1 & 1.8 & 4.0 & 9.3 & 13.1 & 13.3 \\
PANDAS (Ours) & \textbf{56.9} & 57.3 & 57.9 & 58.4 & 58.8 & 58.3 & 58.3 & \textbf{5.7} & \textbf{12.4} & \textbf{14.1} & \textbf{18.6} & \textbf{21.3} & \textbf{22.2} & \textbf{22.2} \\
\bottomrule
\end{tabular}
}
\end{table*}

% ------------------------------------------------
\myparagraph{Replacing Hungarian Matching with vision-language-based evaluation}~Since PANDAS performs NCD in an unsupervised way, we must map its discovered clusters to semantic class labels to evaluate its performance. Following standard NCD evaluation protocols~\citep{Han2020Automatically,fini2021unified,weng2021unsupervised} and prior work on NCD for detection~\citep{Fomenko_2022_NeurIPS}, we used the Hungarian Matching algorithm for our main evaluations. However, this requires an annotated dataset of novel classes to be provided, which might not always be feasible to obtain in practical scenarios. One way to avoid this need for annotated data would be to use a pre-trained vision-language model to map the discovered clusters to classes.

More specifically, in this experiment, we explore the use of CLIP (ViT-B/32)~\citep{radford2021learning} for assigning semantic class labels to discovered clusters. To do this, we first determine the assignment of all box proposals in the discovery set to the clusters discovered by PANDAS. We then embed these box proposals in CLIP feature space. Next, we create a class dictionary from the class labels in VOC and embed each of these labels in CLIP feature space using prompts of the form ``a photo of a \{\texttt{class}\}''. To assign a semantic class label to each cluster, we then determine the closest text embedding to each box proposal embedding in feature space. Given the cluster assignments of the box proposals, we then determine the class labels assigned to a particular cluster. Specifically, we assign a class label to a cluster by taking the statistical mode of the top-$\kappa$ closest box proposal features to the cluster centroid. We then use these cluster-to-class label assignments to compute model performance. Note, for base prototypes, we use the ground truth class label since it is known.

We report the results from this experiment in Table~\ref{tab:voc-clip-eval}, where we compare the performance of PANDAS using Hungarian Matching to the CLIP-based mapping procedure using top-$\kappa$ values in \{1, 5, 10, 25, 50, 100\}. We use the 250 novel cluster version of PANDAS, which we used for our main experiments. Overall, we find that performance on novel classes is worse when using the CLIP mapping as compared to using the Hungarian Matching algorithm. This is not surprising as Hungarian Matching has access to annotated boxes of novel classes. Moreover, we find that using a $\kappa$ value greater than one improves performance on the novel classes, but $\kappa$ values of 25, 50, and 100 yield roughly the same performance. Beyond access to ground truth annotations, another possible explanation for the performance gaps is that the Hungarian Matching algorithm maps each class to a single cluster, while the CLIP-based evaluation can potentially map each class to zero or more clusters. Overall, these results indicate that vision-language models like CLIP could be used for semantic class assignment in the future, eliminating the need to annotate a dataset of novel classes. Moreover, a more diverse dictionary could be studied by collecting common class concepts from a large dataset such as LAION-400M~\citep{schuhmann2021laion}. We believe studying vision-language models for mapping discovered clusters to novel classes is an interesting area for future research.

\begin{table}[t]
\caption{\textbf{Evaluating PANDAS using CLIP on VOC.} We compare the use of the Hungarian Matching algorithm for assigning semantic class names to clusters to a mapping method based on the CLIP vision-language model. We study the use of CLIP with several different top-$\kappa$ values. Each mAP value is the average of three runs with different seeds.
\label{tab:voc-clip-eval}}
\centering
\begin{tabular}{lccc}
\toprule
\textsc{Evaluation}& mAP$_{base}$& mAP$_{novel}$& mAP$_{all}$ \\ 
\midrule
CLIP ($\kappa=1$)& 56.8 & 11.9 & 34.3 \\
CLIP ($\kappa=5$)& 56.6 & 12.0 & 34.3 \\
CLIP ($\kappa=10$)& 56.7 & 12.8 & 34.8 \\
CLIP ($\kappa=25$)& 56.7 & 13.5 & 35.1 \\
CLIP ($\kappa=50$)& 56.7 & 13.6 & 35.1 \\
CLIP ($\kappa=100$)& 56.8 & 13.4 & 35.1 \\
\midrule
Hungarian Matching & \textbf{58.8} & \textbf{21.3} & \textbf{40.0} \\
\bottomrule
\end{tabular}
\end{table}

% ------------------------------------------------
\myparagraph{Evaluation with class-agnostic metrics}~In our paper, the primary performance metric we use is mean average precision (mAP) over classes. While this metric is useful for understanding the holistic behavior of models, it can often hide more fine-grained sources of errors. To study these error sources in more detail, we compute the performance of PANDAS using four additional precision metrics that are class-agnostic. The first evaluates how well the model predicts boxes with any class label. The second evaluates how well the model predicts novel class boxes and assigns them a novel class label. The last two metrics evaluate model errors. That is, we evaluate how often the model predicts a novel class label when the ground truth box belongs to a base class (i.e., the prediction is a novel class, but the ground truth label is a base class), and vice versa (i.e., the prediction is a base class, but the ground truth label is a novel class). The first two metrics demonstrate the difficulty in detecting the presence of unknown objects. This problem is actively studied in literature as \emph{unsupervised object discovery} (see \citet{simeoni2023unsupervised} for a survey). The last two metrics help us better understand model confusion between base and novel classes.

We computed these metrics using the PANDAS models from our main paper that were trained on VOC using three different seeds and computed using 250 novel clusters. We report the average performance over these runs using the four class-agnostic metrics next. Overall, we find that it is challenging for PANDAS to detect the presence of objects (39.7\% AP), but it is especially difficult for it to detect novel boxes as novel classes (15.5\% AP). Surprisingly, we find that PANDAS does not often mis-assign base boxes as novel classes (5.4\% AP) or novel boxes as base classes (10.3\% AP). Based on these results, it seems that one of the biggest challenges for PANDAS is discovering boxes of novel objects. We believe this is an important area of active research (i.e., unsupervised object discovery) that could be combined with novel class discovery methods in the future.

\begin{figure}[t!]
    \centering
        \includegraphics[height=0.20\linewidth]{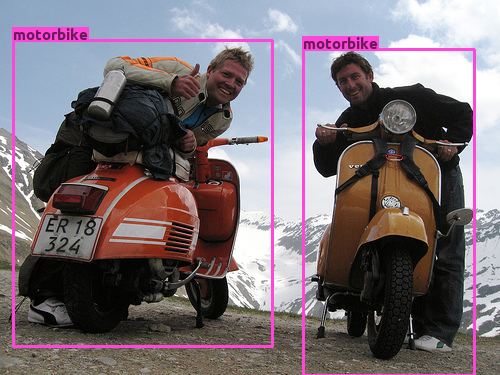}
        \includegraphics[height=0.20\linewidth]{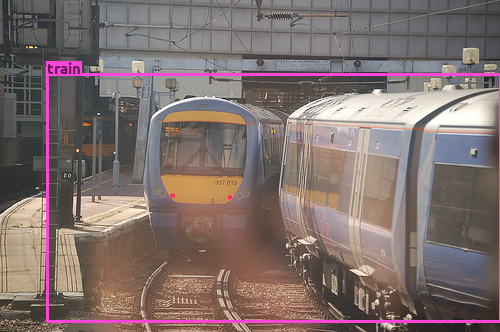}
        \includegraphics[height=0.20\linewidth]{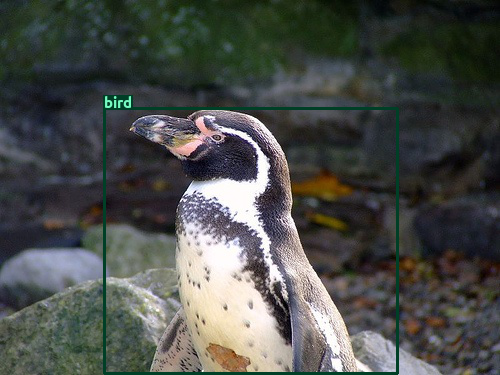}
        \includegraphics[height=0.20\linewidth]{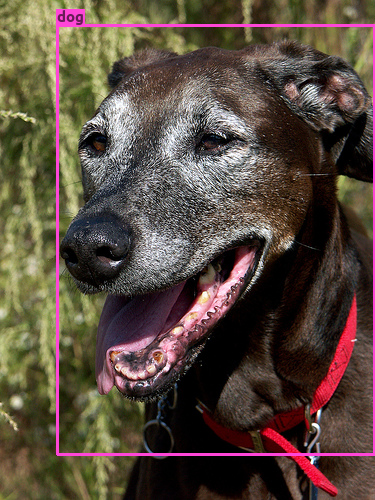}
%    \\
    \caption{\textbf{Cases where PANDAS successfully detects vehicles or centered animals.} 
    Detections output by PANDAS on VOC. We show the \basecolor{Base} classes and \novelcolor{Novel} classes.
    }
    \label{fig:voc-qual-add1}
\end{figure}

\begin{figure}[t!]
    \centering
        \includegraphics[height=0.23\linewidth]{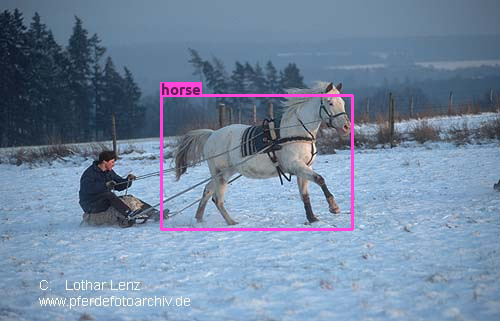}
        \includegraphics[height=0.23\linewidth]{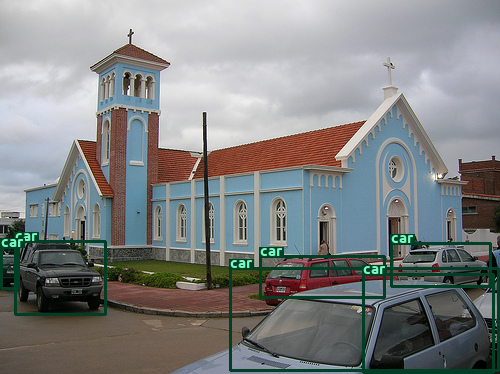}
        \includegraphics[height=0.23\linewidth]{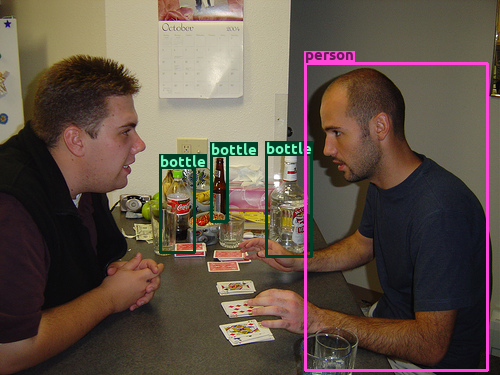}
%    \\
    \caption{\textbf{Cases where PANDAS misses the ``person'' class.} 
    Detections output by PANDAS on VOC. We show the \basecolor{Base} classes and \novelcolor{Novel} classes.
    }
    \label{fig:voc-qual-add2}
\end{figure}

\begin{figure}[t!]
    \centering
        \includegraphics[height=0.19\linewidth]{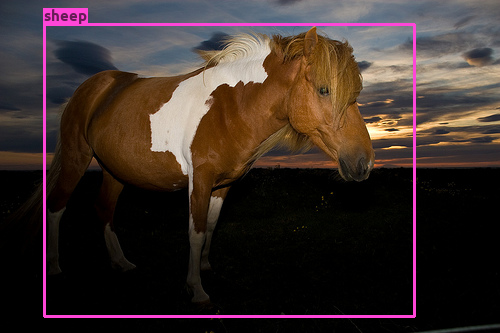}
        \includegraphics[height=0.19\linewidth]{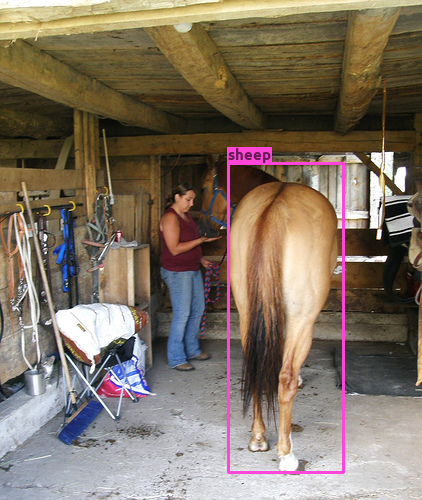}
        \includegraphics[height=0.19\linewidth]{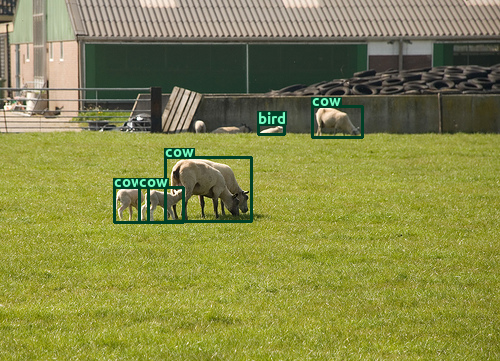}
        \includegraphics[height=0.19\linewidth]{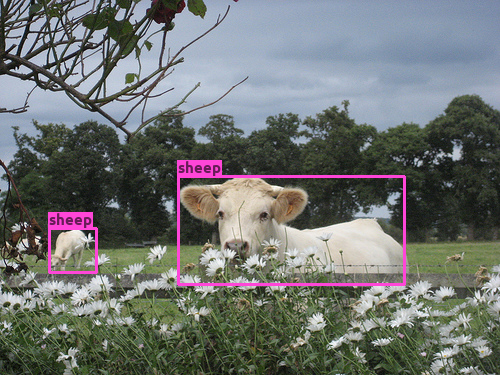}
%    \\
    \caption{\textbf{Cases where PANDAS struggles with semantically similar animal classes.} 
    Detections output by PANDAS on VOC. We show the \basecolor{Base} classes and \novelcolor{Novel} classes.
    }
    \label{fig:voc-qual-add3}
\end{figure}

% ------------------------------------------------
\myparagraph{Additional qualitative examples}~We provide additional examples of the predictions from PANDAS in Fig.~\ref{fig:voc-qual-add1}-\ref{fig:voc-qual-add3}. Fig.~\ref{fig:voc-qual-add1} demonstrates that PANDAS can successfully detect vehicles and centered animals. There is some confusion in disambiguating the two trains in the second image, but given that this is a novel class, it is not very surprising. In Fig.~\ref{fig:voc-qual-add2}, we demonstrate cases where PANDAS struggles to detect the novel ``person'' class, even though it successfully detects other objects (including a person in the third image). In Fig.~\ref{fig:voc-qual-add3}, we demonstrate that semantically similar animal classes are challenging for PANDAS. Specifically, PANDAS confuses the ``horse'' and ``sheep'' classes, as well as ``cow'' and ``sheep''.

% ------------------------------------------------
\section{Additional Results on COCO-to-LVIS}
\label{supp:results-map-0.5-0.95}

\begin{table*}[t]
\caption{Performance on the \COCOtoLVIS benchmark with 1000, 3000, and 5000 novel clusters. We compare PANDAS to RNCDL with a uniform prior (Uniform) and the default RNCDL configuration with a long-tailed prior. We report mAP at an IoU of 0.5 for base, novel, all, frequent (f), common (c), and rare (r) classes. For PANDAS, we report the mean and standard deviation over three runs.
\label{tab:lvis-num-clusters-0.5}}
\centering
\small
\begin{tabular}{lccccccc}
\toprule
\textsc{Method} & \textsc{Clusters} & mAP$_{base}$ & mAP$_{novel}$ & mAP$_{all}$ & mAP$_{f}$ & mAP$_{c}$ & mAP$_{r}$ \\
\midrule
RNCDL (Uniform) & 1000 & 36.2 & 4.8 & 7.2 & 11.0 & 5.2 & 3.6 \\
RNCDL (Uniform) & 3000 & 39.4 & \textbf{5.5} & \textbf{8.1} & 11.4 & \textbf{6.2} & 5.2 \\
RNCDL (Uniform) & 5000 & \textbf{41.5} & 5.0 & 7.8 & \textbf{11.6} & 4.9 & \textbf{6.5} \\
\midrule
RNCDL & 1000 & 36.3 & 5.0 & 7.4 & 11.0 & 5.6 & 3.8 \\
RNCDL & 3000 & 39.9 & \textbf{6.2} & \textbf{8.9} & \textbf{11.7} & \textbf{7.1} & 6.5 \\
RNCDL & 5000 & \textbf{40.8} & 6.1 & 8.8 & \textbf{11.7} & 6.4 & \textbf{8.1} \\
\midrule
PANDAS (Ours) & 1000 & 44.9\footnotesize{{\color{DarkGray}$\pm$0.3}} & 4.3\footnotesize{{\color{DarkGray}$\pm$0.1}} & 7.4\footnotesize{{\color{DarkGray}$\pm$0.1}} & 13.0\footnotesize{{\color{DarkGray}$\pm$0.2}} & 4.1\footnotesize{{\color{DarkGray}$\pm$0.2}} & 2.9\footnotesize{{\color{DarkGray}$\pm$0.4}} \\
PANDAS (Ours) & 3000 & 45.3\footnotesize{{\color{DarkGray}$\pm$0.1}} & 7.1\footnotesize{{\color{DarkGray}$\pm$0.2}} & 10.0\footnotesize{{\color{DarkGray}$\pm$0.2}} & 14.3\footnotesize{{\color{DarkGray}$\pm$0.1}} & 7.3\footnotesize{{\color{DarkGray}$\pm$0.3}} & 7.0\footnotesize{{\color{DarkGray}$\pm$1.4}} \\
PANDAS (Ours) & 5000 & \textbf{45.6}\footnotesize{{\color{DarkGray}$\pm$0.2}} & \textbf{8.7}\footnotesize{{\color{DarkGray}$\pm$0.5}} & \textbf{11.5}\footnotesize{{\color{DarkGray}$\pm$0.4}} & \textbf{14.7}\footnotesize{{\color{DarkGray}$\pm$0.1}} & \textbf{9.2}\footnotesize{{\color{DarkGray}$\pm$0.6}} & \textbf{10.4}\footnotesize{{\color{DarkGray}$\pm$1.5}} \\
\bottomrule
\end{tabular}
\end{table*}

% ------------------------------------------------
\myparagraph{Impact of the number of clusters}~We study the impact of the number of novel clusters on RNCDL and PANDAS on the \COCOtoLVIS benchmark using mAP at an IoU of 0.5 in Table~\ref{tab:lvis-num-clusters-0.5}. Following \citet{Fomenko_2022_NeurIPS}, we study performance of RNCDL and PANDAS for 1000, 3000, and 5000 novel clusters. In our main paper, we report performance for RNCDL with 3000 novel clusters, as suggested in the original paper~\citep{Fomenko_2022_NeurIPS}. For PANDAS, we report performance in our main paper with 5000 novel clusters. In this experiment, we use the variant of RNCDL without mask annotations, which is more comparable to PANDAS.

The overall performance (mAP$_{all}$) for RNCDL (Uniform) is best when 3000 novel clusters are used; however, performance of the long-tailed variant of RNCDL is consistently the same as or better than the uniform variant. This is expected as LVIS is extremely long-tailed and assuming a long-tailed prior on class labels is helpful to the model. For the long-tailed variant of RNCDL, overall performance (mAP$_{all}$) is best when using 3000 novel clusters, which is consistent with \citet{Fomenko_2022_NeurIPS}. However, for 5000 novel clusters, RNCDL exhibits slightly better performance on base (+0.9\%) and rare (+1.6\%) classes. For PANDAS, using 5000 novel clusters is consistently the best, across all metrics.

Moreover, these results also indicate that the number of clusters is an 
important hyper-parameter in NCD, which requires careful consideration. In this paper, we made the effort of comparing PANDAS (our proposed method) with RNCDL~\citep{Fomenko_2022_NeurIPS} (the state of the art) with different hyper-parameter settings and design choices.

\begin{figure}[h!]
        \centering
        \includegraphics[height=0.26\linewidth]{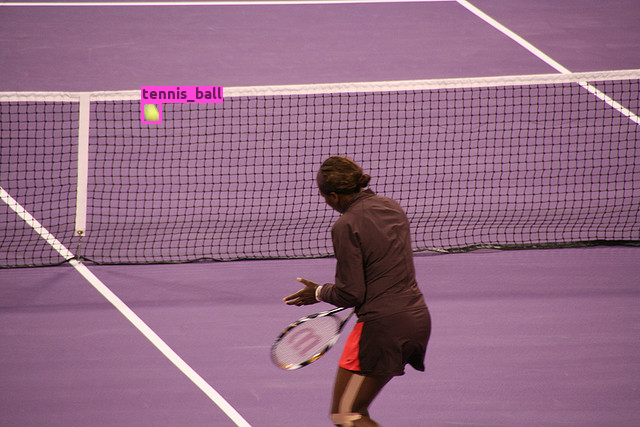}
        \includegraphics[height=0.26\linewidth]{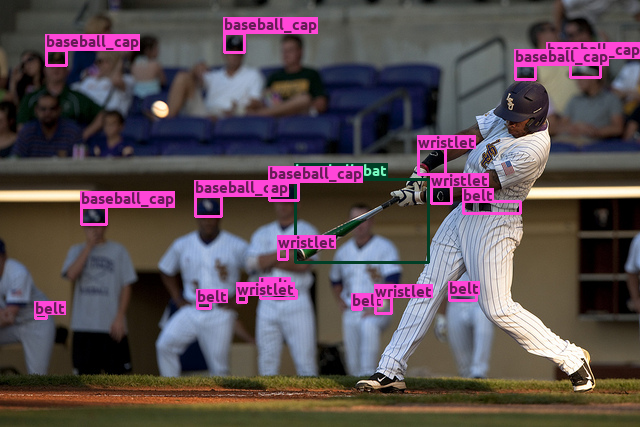}
        \includegraphics[height=0.26\linewidth]{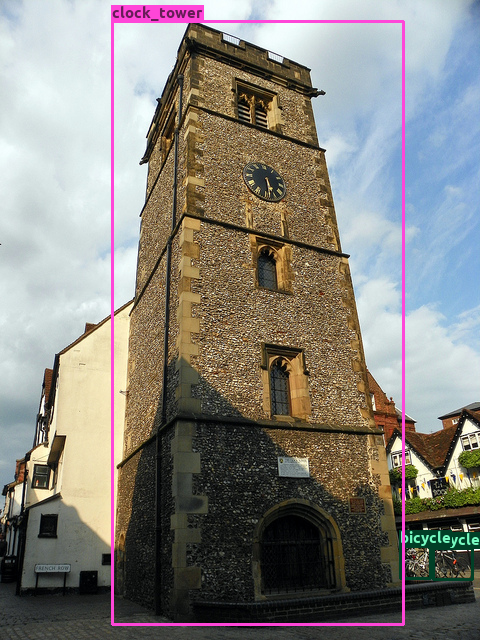}
        \\
        \includegraphics[height=0.26\linewidth]{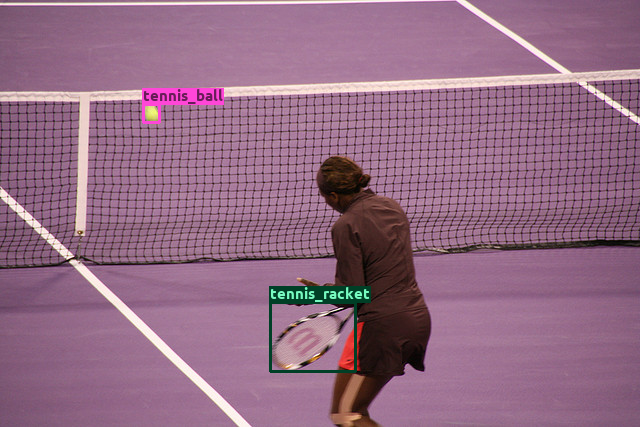}
        \includegraphics[height=0.26\linewidth]{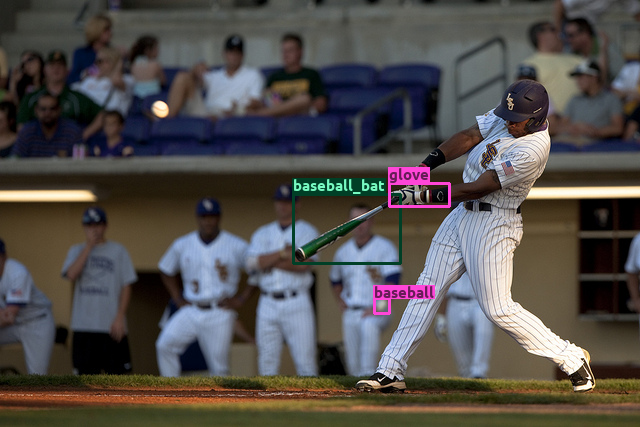}
        \includegraphics[height=0.26\linewidth]{lvis_qual_images/viz_lvis_245173_predictions_iou_nms_5000_clusters_ha_2_color.png}
    \caption{\textbf{Cases where PANDAS successfully detects objects.} Ground truth (top) and PANDAS outputs (bottom) with \basecolor{Base} and \novelcolor{Novel} classes on \COCOtoLVIS.
    }
    \label{fig:lvis-qual-add1}
\end{figure}

\begin{figure}[h!]
        \centering
        \includegraphics[height=0.225\linewidth]{lvis_qual_images_supp/viz_lvis_24813_gt_5000_clusters_ha_2_color.png}
        \includegraphics[height=0.225\linewidth]{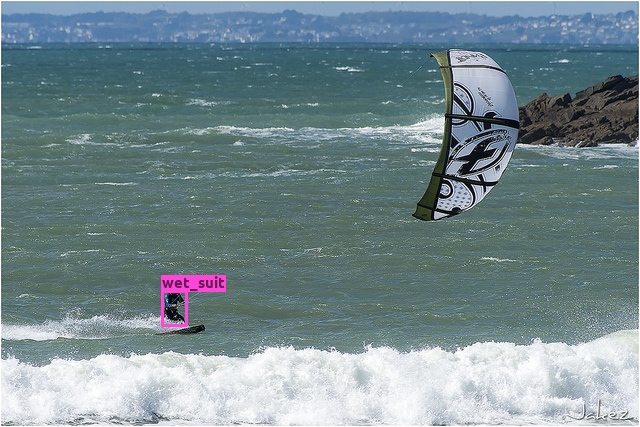}
        \includegraphics[height=0.225\linewidth]{lvis_qual_images_supp/viz_lvis_440311_gt_5000_clusters_ha_2_color.png}
        \\
        \includegraphics[height=0.225\linewidth]{lvis_qual_images_supp/viz_lvis_24813_predictions_iou_nms_5000_clusters_ha_2_color.png}
        \includegraphics[height=0.225\linewidth]{lvis_qual_images_supp/viz_lvis_60069_predictions_iou_nms_5000_clusters_ha_2_color.png}
        \includegraphics[height=0.225\linewidth]{lvis_qual_images_supp/viz_lvis_440311_predictions_iou_nms_5000_clusters_ha_2_color.png}
    \caption{\textbf{Cases where PANDAS outputs reasonable predictions that are not included in ground truth.} Ground truth (top) and PANDAS outputs (bottom) with \basecolor{Base} and \novelcolor{Novel} classes on \COCOtoLVIS.
    }
    \label{fig:lvis-qual-add2}
\end{figure}

\begin{figure}[ht!]
        \centering
        \includegraphics[height=0.23\linewidth]{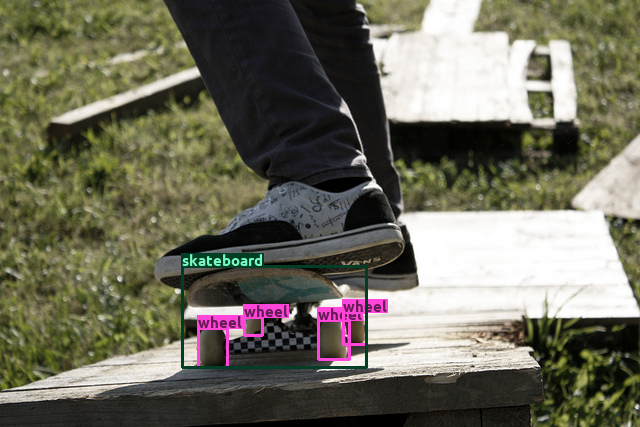}
        \includegraphics[height=0.23\linewidth]{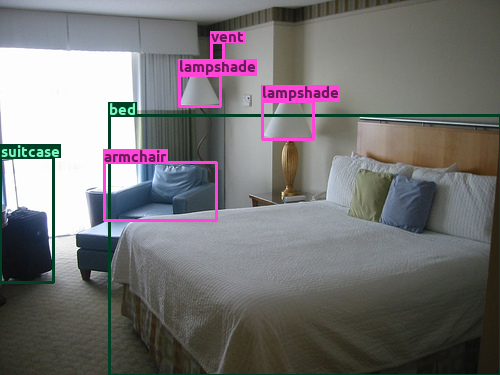}
        \includegraphics[height=0.23\linewidth]{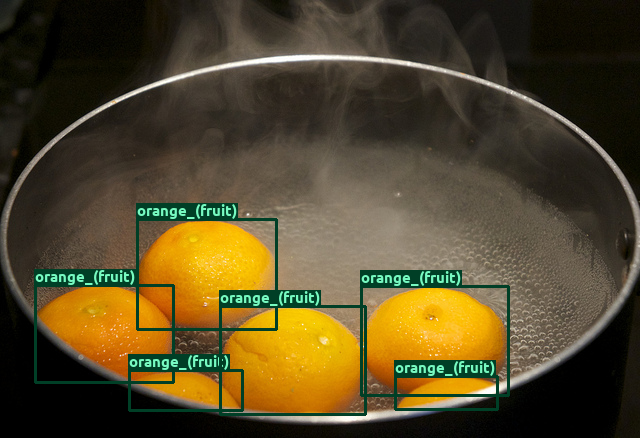}
        \\
        \includegraphics[height=0.23\linewidth]{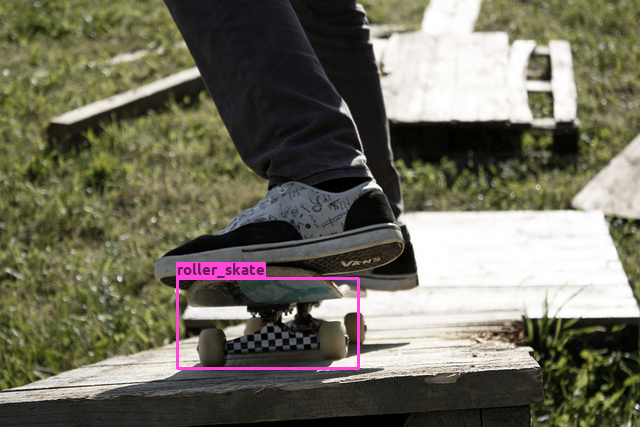}
        \includegraphics[height=0.23\linewidth]{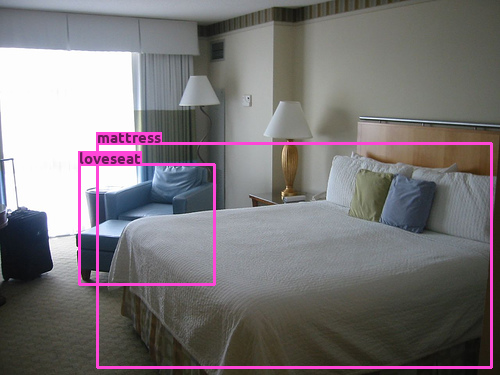}
        \includegraphics[height=0.23\linewidth]{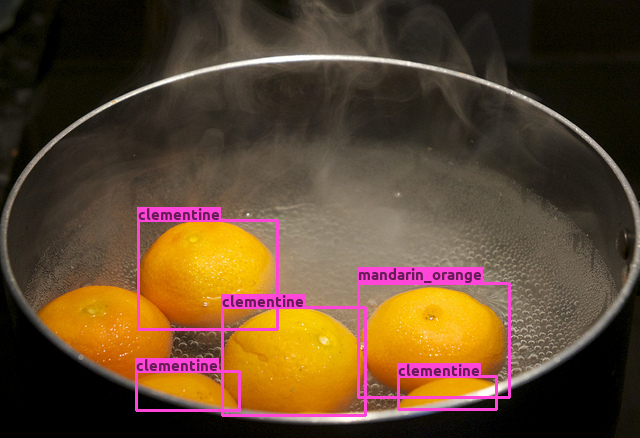}
        \\
        \includegraphics[height=0.23\linewidth]{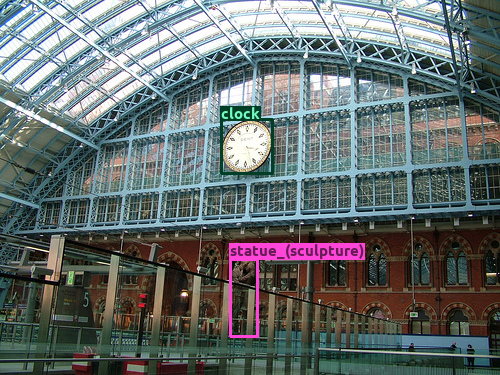}
        \includegraphics[height=0.23\linewidth]{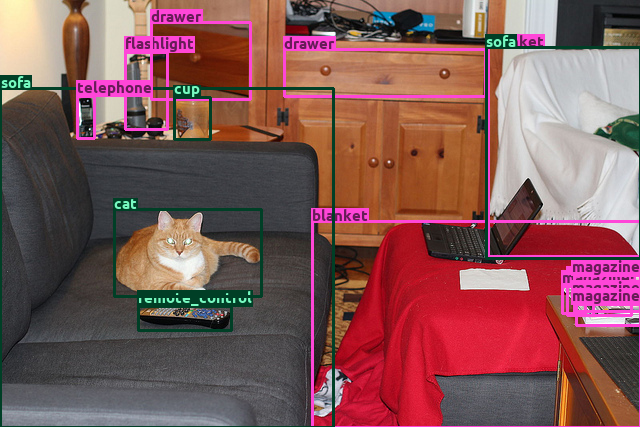}
        \includegraphics[height=0.23\linewidth]{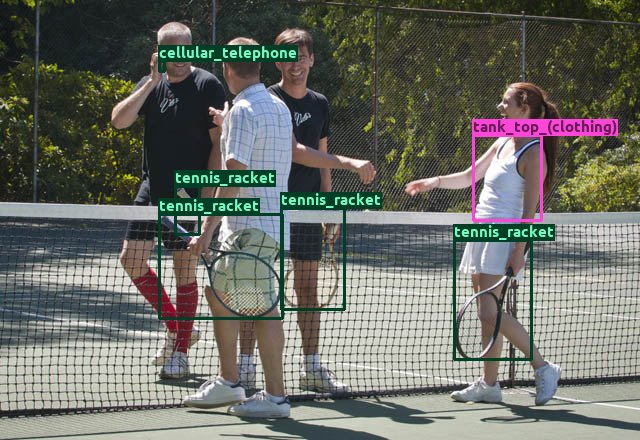}
        \\
        \includegraphics[height=0.23\linewidth]{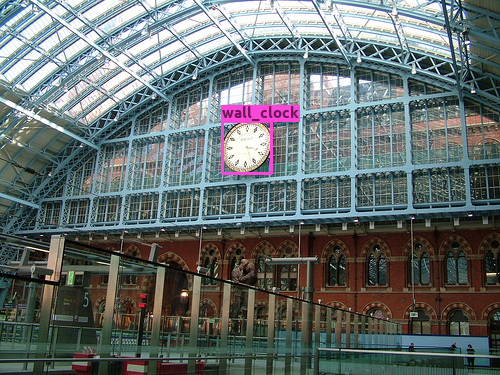}
        \includegraphics[height=0.23\linewidth]{lvis_qual_images_supp/viz_lvis_271728_predictions_iou_nms_5000_clusters_ha_2_color.png}
        \includegraphics[height=0.23\linewidth]{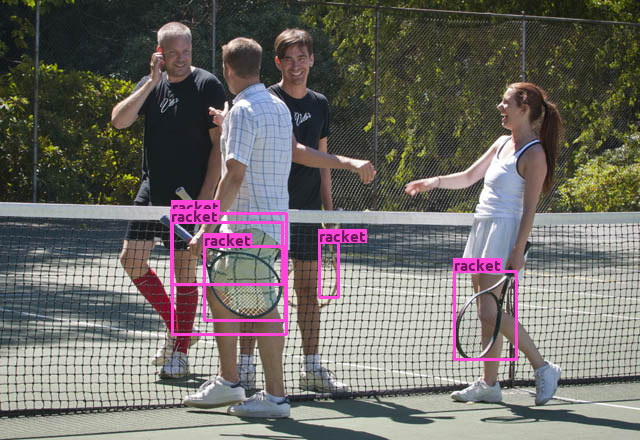}
    \caption{\textbf{Cases where PANDAS faces semantically similar classes.} Ground truth (row 1 and 3) and the associated PANDAS outputs (row 2 and 4) with \basecolor{Base} and \novelcolor{Novel} classes on \COCOtoLVIS.
    }
    \label{fig:lvis-qual-add3}
\end{figure}

% ------------------------------------------------
\myparagraph{Additional qualitative examples}~We provide additional qualitative examples of the predictions produced by PANDAS on \COCOtoLVIS in Fig.~\ref{fig:lvis-qual-add1}-\ref{fig:lvis-qual-add3}. In Fig.~\ref{fig:lvis-qual-add1}, we show successful detections from PANDAS, although we find that it sometimes struggles to identify many novel classes (e.g., the second image). In Fig.~\ref{fig:lvis-qual-add2}, we show cases where PANDAS outputs reasonable predictions that are not included in the ground truth annotations. These examples showcase the importance of qualitative inspections for NCD, as ground truth annotations or class assignments from the Hungarian Matching algorithm could be inaccurate or lacking. In Fig.~\ref{fig:lvis-qual-add3}, we show cases where PANDAS faces the challenges of semantic similarity among classes. We find that some of the outputs from PANDAS are quite reasonable (e.g., ``bed'' vs. ``mattress'', ``clock'' vs. ``wall clock'', ``cat'' vs. ``pet'', and ``tennis racket'' vs. ``racket''). In addition to quantitative metrics, we encourage future researchers to inspect qualitative outputs from NCD models since quantitative metrics may not reveal all successes or failures of a model.

\clearpage

\end{document}